\pdfoutput=1
\documentclass[11pt]{article}

\usepackage[final]{acl}
\usepackage{times}
\usepackage{latexsym}
\usepackage[T1]{fontenc}
\usepackage[utf8]{inputenc}
\usepackage{microtype}
\usepackage{inconsolata}

\usepackage{tikz}
\usepackage{algpseudocode}
\algrenewcommand\textproc{\text}
\usepackage{makecell}
\usepackage{booktabs}
\usepackage{color}
\usepackage{colortbl}
\usepackage{multirow}
\usepackage{enumitem}
\usepackage{makecell}
\usepackage{threeparttable}
\usepackage{amsmath,amsfonts,mathtools}%,amssymb}
\usepackage{pifont}

\usepackage{mathrsfs}
\usepackage{float}
\usepackage{graphicx}
\usepackage{xcolor}
\usepackage{enumitem}
\usepackage{tabularx, booktabs}

\usepackage[most]{tcolorbox}
\usepackage{fvextra}

\definecolor{mygreen}{HTML}{00B050}

\definecolor{myorange}{HTML}{ED7D31}
\definecolor{rowgray}{gray}{0.97}
\definecolor{avgblue}{RGB}{210,230,250}  % 浅蓝色（接近表格中浅蓝效果）
\definecolor{headergray}{RGB}{160,160,160} % 深灰色（可根据视觉微调RGB值，示例为常见深灰）

\usepackage{array}
\usepackage{mathtools}
\usepackage{multirow}
\usepackage{subcaption}
\usepackage{tikz}
\definecolor{color1}{cmyk}{0.216,0.176,0,0}
\definecolor{color2}{cmyk}{0.059,0.235,0.392,0}

\usepackage{graphicx}
\usepackage{tikz}
\usepackage{wrapfig}
\usepackage{algorithm}
\usepackage{algpseudocode}
\algrenewcommand\textproc{\text}
\usepackage{makecell}
\usepackage{booktabs}
\usepackage{pifont}
\usepackage{multirow}
\usepackage{enumitem}
\usepackage{balance}
\usepackage{threeparttable}
\usepackage{amsmath,amsfonts,mathtools} % 已经包含了 amssymb 的大部分功能
\usepackage{longtable}%yzb在后面加的库
\usepackage{amsthm}

\usepackage{colortbl}
\usepackage{mathrsfs}
\usepackage{float}
\usepackage{graphicx}
\usepackage{hyperref}
\usepackage{tabularx, booktabs}
\usepackage{tikz}
\usepackage{CJKutf8}

\definecolor{uc_color}{rgb}{0.99,0.24,0.63}
\definecolor{hc_color}{rgb}{0.02,0.51,0.51}
\definecolor{tc_color}{rgb}{0.99,0.55,0.09}

\tikzstyle{mybox} = [draw=black, very thick,
    rectangle, rounded corners, inner sep=10pt, inner ysep=13pt]
\tikzstyle{fancytitle} =[fill=black, text=white]

\newcommand{\mname}{\textsc{StackPlanner}}

\title{\mname: A Centralized Hierarchical Multi-Agent System with Task-Experience Memory Management}

\author{Ruizhe Zhang$^{1,2,3}$\thanks{Ruizhe, Xinke, Zhibang, Zhixin, and Jiaran contributed equally to this work. Ruizhe and Xinke led the design and implementation of the multi-agent interaction mechanism in \mname. Xinke, Zhixin, and Jiaran developed the short-term memory stack framework and training mechanism. Xinke and Zhixin developed the long-term memory mechanism. Zhibang, Yuzhen, Zhengxing and Yuxuan developed the report generation component. Zhibang, Yuzhen and Yuxuan developed the multi-turn dialogue and human-AI interaction components.}, Xinke Jiang$^{1,2,3}$\footnotemark[1], Zhibang Yang$^{1,2}$\footnotemark[1], Zhixin Zhang$^{1,2,3}$\footnotemark[1], Jiaran Gao$^{1,2}$\footnotemark[1] \\
\textbf{Yuzhen Xiao}$^{1,2,3}$\textbf{,}
\textbf{Tao Feng}$^{4}$\textbf{,}
\textbf{Yue Fang}$^{1,2}$\textbf{,}
\textbf{Yuxuan Liu}$^{2}$\textbf{,}
\textbf{Ruiqing Li}$^{1,2}$\textbf{,}
\textbf{Hongbin Lai}$^{5}$\textbf{,} 
\textbf{Huheng Huang}$^{6}$\textbf{,} 
\\
\textbf{Xu Chu}$^{1,3,7}$\footnotemark[2]\textbf{,} \textbf{Junfeng Zhao}$^{1,3}$\thanks{Corresponding author.}\textbf{,} \textbf{Yasha Wang}$^{2,8}$\footnotemark[2]\\
\textsuperscript{1}School of Computer Science, Peking University, Beijing, China\\[-0.1em]
\textsuperscript{2}National Engineering Research Center for Software Engineering, Peking University, Beijing, China\\[-0.1em]
\textsuperscript{3}Key Laboratory of High Confidence Software Technologies, Ministry of Education, Beijing, China\\[-0.1em]
\textsuperscript{4}School of Public Affaris, Zhejiang University, Hangzhou, China\\[-0.1em]
\textsuperscript{5}School of Software \& Microelectronics, Peking University, Beijing, China\\[-0.1em]
\textsuperscript{6}GRG Banking Equipment Co., Ltd., Guangzhou, China\\[-0.1em]
\textsuperscript{7}Center on Frontiers of Computing Studies, Peking University, Beijing, China\\[-0.1em]
\textsuperscript{8}Peking University Information Technology Institute (Tianjin Binhai), Tianjin, China\\[-0.1em]
    \small
    \{nostradamus, xinkejiang, yangzb\}@stu.pku.edu.cn, \{chu\_xu, zhaojf, wangyasha\}@pku.edu.cn
}

\begin{document}
\maketitle
\begin{abstract}
Multi-agent systems based on large language models, particularly centralized architectures, have recently shown strong potential for complex and knowledge-intensive tasks. However, central agents often suffer from unstable long-horizon collaboration due to the lack of memory management, leading to context bloat, error accumulation, and poor cross-task generalization. To address both task-level memory inefficiency and the inability to reuse coordination experience, we propose \mname~, a hierarchical multi-agent framework with explicit memory control. \mname~ addresses these challenges by decoupling high-level coordination from subtask execution with active task-level memory control, and by learning to retrieve and exploit reusable coordination experience via structured experience memory and reinforcement learning. Experiments on multiple deep-search and agent system benchmarks demonstrate our effectiveness in enabling reliable long-horizon multi-agent collaboration.
\end{abstract}

\section{Introduction}
\begin{figure}[t]
  \centering                                                                                         
  \includegraphics[width=0.48\textwidth]{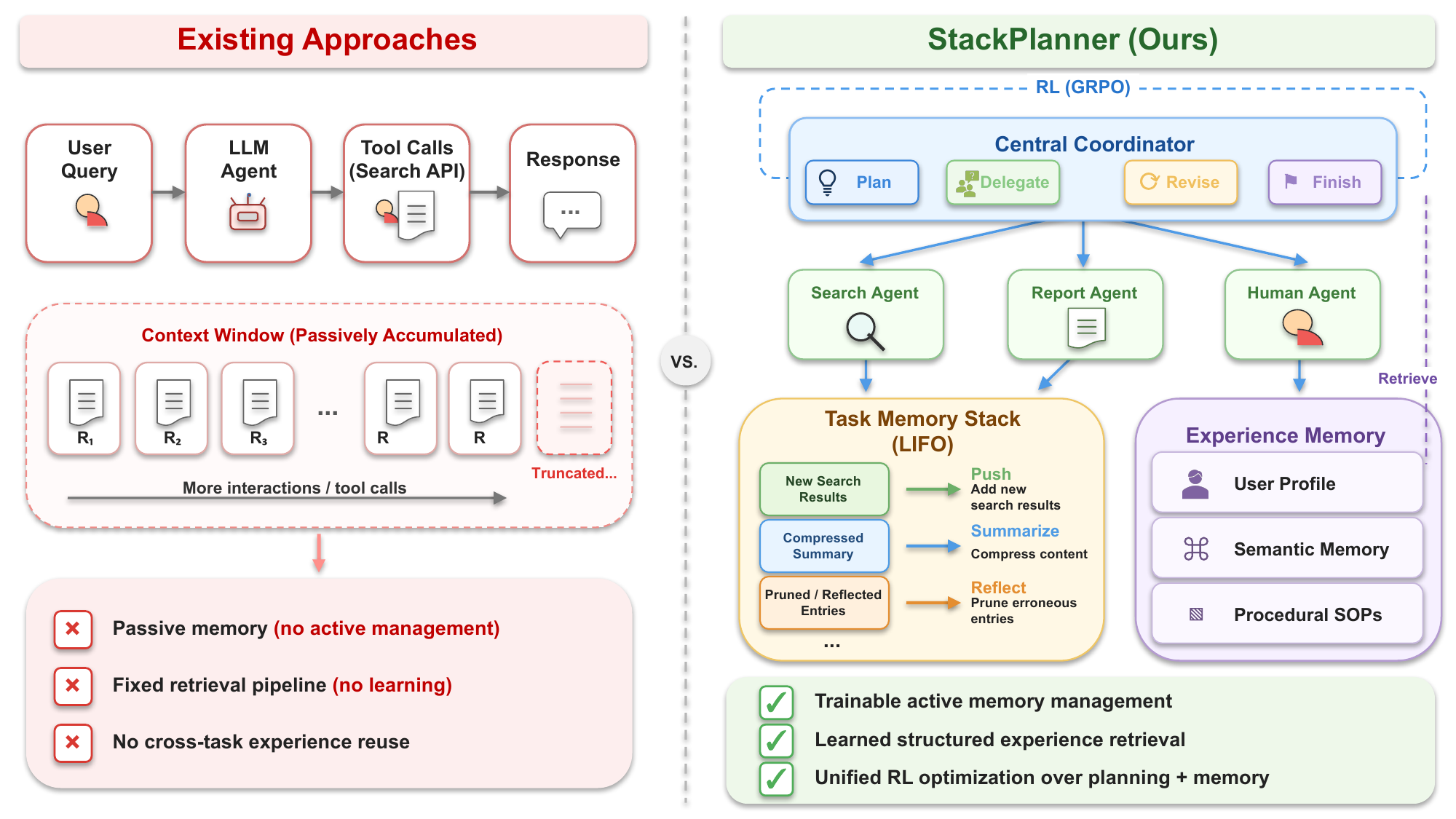}
  \caption{Comparison of existing approaches and \mname{}. (a)~Existing multi-agent systems treat memory as a passive byproduct with fixed summarization or truncation strategies. (b)~\mname{} introduces trainable active memory management: the coordinator learns \emph{when} and \emph{how} to revise task memory via RL-optimized \textsc{Revise} actions, and retrieves structured experience memory to guide planning and delegation.}
  \label{fig:motivation} 
\end{figure}  

Large Language Model-based multi-agent systems (LLM-MAS) have emerged as an effective paradigm for addressing complex, long-horizon, and knowledge-intensive tasks~\cite{chen2025surveyllmbasedmultiagentsystem, guo2024largelanguagemodelbased}. By enabling task decomposition, parallel exploration, and collaborative reasoning, these systems have been applied to challenging problem-solving and information-intensive scenarios~\cite{wu2023autogenenablingnextgenllm, hong2024metagptmetaprogrammingmultiagent, qian2024chatdevcommunicativeagentssoftware}. Prior work has explored a variety of designs, including decentralized collaboration~\cite{yang2025agentnetdecentralizedevolutionarycoordination}, debate-based collectives~\cite{du2023improvingfactualityreasoninglanguage}, and structured multi-stage reasoning pipelines~\cite{yao2023treethoughtsdeliberateproblem}.  
However, as system scale and task complexity increase, ensuring reliable multi-agent collaboration over long-horizon, information-intensive, and cross-task scenarios remains a central dilemma~\cite{guo2024largelanguagemodelbased}. Decentralized and debate-based approaches provide flexibility and robustness but often suffer from high communication overhead, redundant reasoning, and uncertainty in maintaining global consistency~\cite{yang2025agentnetdecentralizedevolutionarycoordination, cui2025freemadconsensusfreemultiagentdebate}. To mitigate these issues, most studies adopt a centralized coordination paradigm, introducing a \textbf{central agent} to unify planning, task allocation, and information integration by operating sub-agents to a unified decision-making framework.~\cite{hou2024coactgloballocalhierarchyautonomous, yue2025masrouterlearningroutellms}. These sub-agents are typically responsible for specialized tasks such as retrieval and knowledge grounding~\cite{HyKGE,asai2023selfrag,zhang2024knowpoknowledgeawarepreferenceoptimization,xu2025parentingoptimizingknowledgeselection} or deep research and report generation~\cite{li2025webweaverstructuringwebscaleevidence}.

Despite its advantages, most centralized multi-agent systems \textbf{place the entire burden of coordination, information integration, and decision-making on a single central agent}. As tasks grow in scale and complexity, the influx of information and long reasoning chains can overwhelm the central agent’s processing capacity~\cite{jiang2024tcrag, liu2023lostmiddlelanguagemodels, liu2024longgenbenchlongcontextgenerationbenchmark}, significantly degrading its performance. This limitation is especially pronounced in novel domains or tasks with little prior experience. 
Crucially, both issues stem from the central agent’s limited \textit{\textbf{memory management}} capabilities, encompassing both task-level and cross-task memory. Addressing this deficiency gives rise to two key challenges:

\noindent\textit{\textbf{\ding{182} Challenge 1. How can the central agent’s task memory be effectively managed to mitigate contextual noise and memory bloat, ensuring stable decision-making over long-horizon tasks?}}  
As tasks unfold, information from multiple sub-agents is often redundant or noisy, yet it is \textbf{indiscriminately appended to the central agent's task memory}. Early errors or noise in sub-tasks or tool invocations can propagate across long-horizon steps, causing the central agent to become \textit{lost in the middle of reasoning}, which may result in plan deviations, imbalanced task allocations, or repeated exploration. Existing methods largely rely on \textit{passive memory management strategies}, such as template-based summarization~\cite{dou-etal-2021-gsum} or heuristic truncation~\cite{liu2023lostmiddlelanguagemodels}, treating memory as a static byproduct rather than a controllable resource. However, without awareness of and active control over its memory state, the central agent’s performance deteriorates as reasoning steps increases.

\noindent\textit{\textbf{\ding{183} Challenge 2. How can valuable historical trajectories (Experience Memory) of the central agent be effectively leveraged to improve task planning and coordination across new tasks?}}  
When tackling new tasks, the central agent often starts from scratch, with \textbf{little reference to prior successful coordination experiences}. Although its decision-making is critical to overall system performance, LLMs are rarely trained for long-horizon, cross-agent reasoning, limiting their ability to \textbf{plan complex tasks effectively}. As a result, systems frequently exhibit poor cold-start performance~\cite{li-etal-2023-theory, li2025fmagent} and limited cross-task generalization~\cite{li2025crosstaskexperientiallearningllmbased}. 

To address these challenges, we construct a \textbf{Hierarchical Multi-Agent System} --- \mname, centered on a coordinator, explicitly supporting the management of \textbf{task memory} and \textbf{experience memory}. 
Unlike prior approaches that treat memory as a passive byproduct, we explicitly model memory as a controllable decision variable and unify task memory management, experience retrieval, and coordination actions within a single reinforcement learning framework.
Specifically:  
\ding{182} For \textbf{\textit{C1}}, we \textbf{decouple the central coordinator’s high-level decision-making from the execution details handled by specialized sub-agents}. By strictly separating the memory of the coordinator and sub-agents , we prevent sub-agents from indiscriminately appending raw execution results to coordinator's task memory, thereby alleviating cognitive and memory pressure on the central agent.
In addition, the central coordinator is equipped with an \textbf{active task memory management mechanism}, enabling it to \textbf{selectively store, condense, and prune task-relevant information}. This mechanism helps mitigate contextual noise and memory bloat, maintain cleaner task representations, and enhance decision-making stability over long-horizon multi-agent interactions.
\ding{183} For \textbf{\textit{C2}}, we introduce a \textbf{experience memory and retrieval module} that stores valuable cross-task coordination experiences, including factual knowledge and procedural memory. This allows the central agent to selectively retrieve relevant historical trajectories, leveraging past strategies and decision patterns to improve planning, delegation, and coordination across new tasks.
To further enhance, we model the full planning process as \textbf{a learnable decision process} and train the coordinator via reinforcement learning.

% Abstract: Experiments on multiple deep-search and deep-research benchmarks demonstrate the effectiveness of our approach in enabling long-horizon multi-agent collaboration.

% In summary, we present a hierarchical multi-agent system that enhances memory management through task-level and long-term memory mechanisms, enabling the central agent to maintain stable decision-making and leverage past experiences for cross-task generalization. 
% Experiments across multiple deep search and deep research benchmarks demonstrate the effectiveness of our approach in long-horizon multi-agent collaboration.

\section{Related Work}

\paragraph{LLM-based multi-agent coordination.}
LLM-based multi-agent systems have become a common paradigm for decomposing complex tasks, assigning specialized roles, and aggregating intermediate results~\cite{guo2024largelanguagemodelbased,chen2025surveyllmbasedmultiagentsystem}. Representative systems such as AutoGen, MetaGPT, and ChatDev show that explicit roles and structured communication can improve tool use, software development, and collaborative problem solving~\cite{wu2023autogenenablingnextgenllm,hong2024metagptmetaprogrammingmultiagent,qian2024chatdevcommunicativeagentssoftware}. Beyond role specialization, recent work explores decentralized or debate-based coordination, where agents exchange proposals or evolve communication policies without a single controller~\cite{du2023improvingfactualityreasoninglanguage,yang2025agentnetdecentralizedevolutionarycoordination,cui2025freemadconsensusfreemultiagentdebate}. These designs improve diversity but can introduce redundant communication and make global state consistency difficult. Centralized coordination therefore remains practical for long-horizon tasks: CoAct introduces a global--local hierarchy, MasRouter learns to route requests among agents, and MacNet, OWL, and AFlow optimize collaboration structures or workflows~\cite{hou2024coactgloballocalhierarchyautonomous,yue2025masrouterlearningroutellms,qianscaling,hu2025owl,zhangaflow}. \mname{} follows this centralized line, but targets the coordinator's memory state: planning, delegation, and memory operations are exposed as one learnable decision process.

\paragraph{Long-horizon agentic search and learning.}
Long-horizon agent systems increasingly combine search, planning, retrieval, and reinforcement learning. Recent search and research agents highlight the value of persistent coordination state: Beyond Ten Turns scales agentic search with asynchronous reinforcement learning, WebWeaver organizes web-scale evidence through dynamic outlines, and Pi-Serini revisits the retrieval backbone for agentic search~\cite{gao2025turnsunlockinglonghorizonagentic,li2025webweaverstructuringwebscaleevidence,hsu2026piserini}. Agent planning and efficiency have also been improved through external knowledge, tree search, or speculative planning~\cite{zhu2024knowagentknowledgeaugmentedplanningllmbased,koh2024treesearchlanguagemodel,choi2026idlespec}. In parallel, RL methods such as Search-R1 and ReSearch train agents to invoke search while reasoning, and Spreadsheet-RL optimizes realistic tool-use trajectories~\cite{jin2025search,chen2025learning,chi2026spreadsheetrl}. These approaches improve exploration, evidence acquisition, or action selection, but usually do not make coordinator memory updates part of the learned action space.

\paragraph{Memory and retrieval in agentic reasoning.}
Long-horizon agents must decide not only what to do next, but also what information should remain available. Existing systems commonly rely on passive context management, including summarization-based compression~\cite{dou-etal-2021-gsum}, heuristic truncation motivated by long-context degradation~\cite{liu2023lostmiddlelanguagemodels}, or retrieval-style context reconstruction~\cite{jiang2024tcrag}. Retrieval-augmented generation learns to select, structure, or critique external evidence, from Self-RAG's reflective retrieval decisions to preference- or tuning-based methods for controllable knowledge selection~\cite{asai2023selfrag,zhang2024knowpoknowledgeawarepreferenceoptimization,xu2025parentingoptimizingknowledgeselection}. These methods ground generation, but mainly operate over external documents rather than the evolving working memory of a multi-agent coordinator. Agent memory has also been studied for reflection, dialogue personalization, and cross-task reuse~\cite{shinn2024reflexion,li2025crosstaskexperientiallearningllmbased,tan-etal-2025-prospect}. However, these approaches usually treat memory as a buffer, retrieved evidence, or post-hoc reflection. \mname{} instead models task memory revision and experience retrieval as first-class coordinator actions, enabling the policy to learn when to condense, prune, or exploit memory jointly with planning and delegation.

\begin{figure*}[t]
\centering
\includegraphics[width=0.98\textwidth]{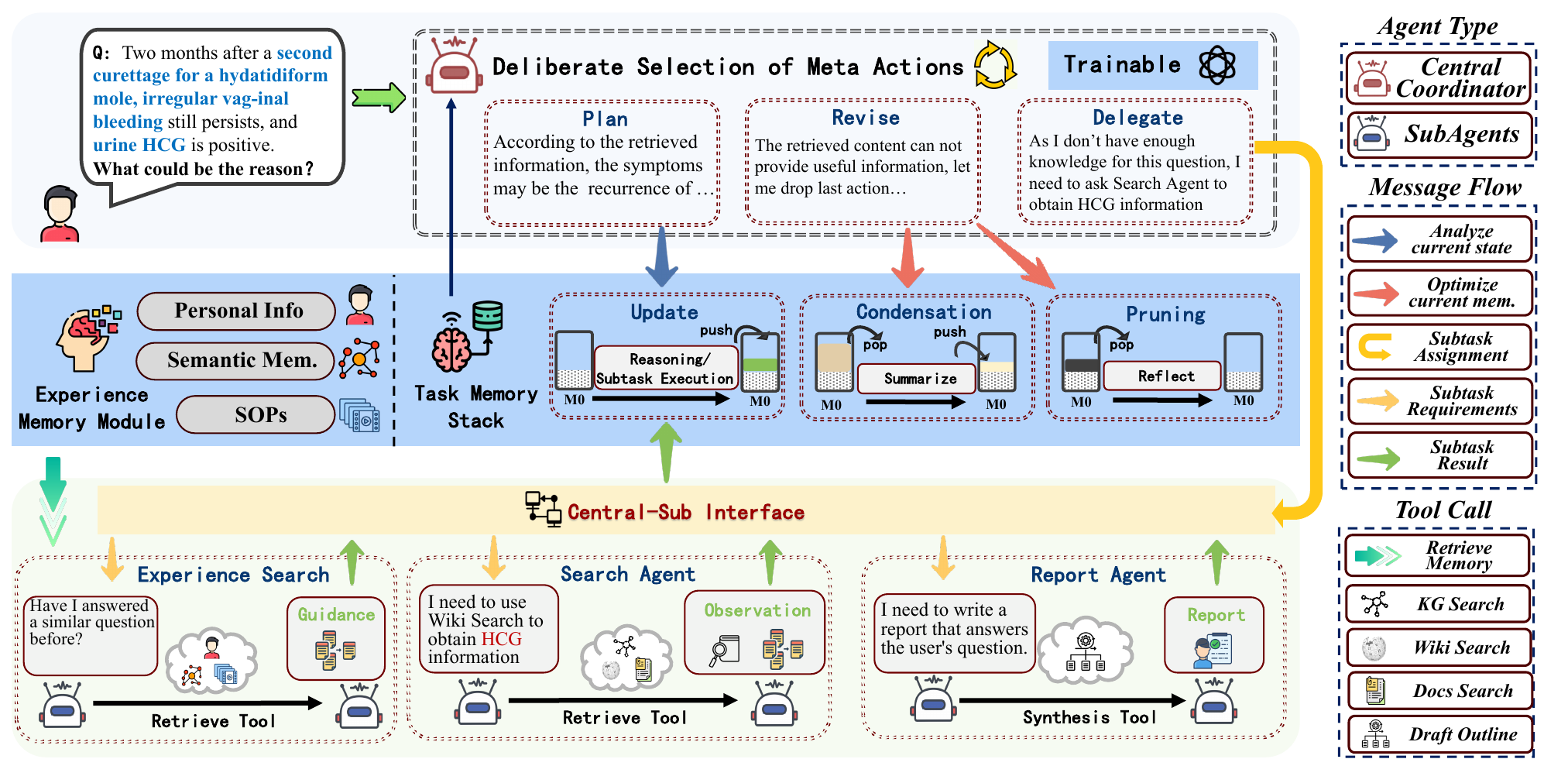}
\caption{Overview of \mname~ framework.}
\label{fig:framework}
\vspace{-0.4cm}
\end{figure*}

\section{Methodology}

As shown in Figure~\ref{fig:framework}, \mname\ follows a hierarchical multi-agent design.
A \textbf{central coordinator} is responsible for high-level decision making,
including planning, subtask delegation, and active memory operations, while specialized sub-agents handle concrete task execution.
Moreover, the coordinator operates over a \emph{task memory} that maintains a concise execution trace,
and leverages a \emph{structured experience memory} that stores reusable knowledge and coordination experience across tasks, which directly address \textbf{\textit{C1}} and \textbf{\textit{C2}}.

\subsection{Hierarchical Coordination}
% \subsection{Hierarchical Coordination[TODO Add Detail]}

\paragraph{Central Coordinator Action Space}

Central coordinator operates over a compact discrete action:
\[
\mathcal{A} = \{\textsc{Plan}, \textsc{Delegate}, \textsc{Revise},\textsc{Finish}\}.
\]
Here \textsc{Plan} determines the next coordination step based on task memory. 
\textsc{Delegate} assigns a scoped subtask to a selected sub-agent, together 
with task requirements and relevant contextual information.
\textsc{Revise} actively optimizes task memory via condensation and pruning, 
and is implemented through two concrete operations, \textsc{Summarize} and \textsc{Reflect}, 
corresponding to memory compression and pruning, respectively.
We additionally include a \textsc{Finish} decision, which is selected by the coordinator to terminate the task when the task is fulfilled.
This action space keeps the coordinator focused on global progress, ensuring system-wide behavior remains task-oriented.
Central coordinator details are in Appendix~\ref{appendix: prompts}.

\paragraph{Specialized Sub-Agents}
Moreover, we also incorporate specialized sub-agents:
\begin{itemize}[leftmargin=*]
    \item \textbf{Search Agent}: conducts information retrieval via tools, 
    following ReAct paradigm for iterative information gathering, organization;
    \item \textbf{Report Agent}: adapts its behavior to the assigned subtask, 
    \emph{either} organizing previous information into structured task reports to 
    support subsequent coordination and execution, \emph{or} invoking 
    professional writing tools to design report structures and populate content for refined text.
\end{itemize}

\subsection{Active Task Memory Management}
% \subsection{Active Task Memory Management [TODO Add Detail]}%增加栈的描述进出的从 TCRAG 之中
Coordinator maintains a lightweight task memory stack
$\mathcal{M} = \{m_1, \dots, m_t\}$, which sequentially stores task execution 
information, and is accessed and modified 
exclusively through \textsc{Revise}.
Task memory stack mechanisms support 3 operations:
\begin{itemize}[leftmargin=*]
    \item \colorbox{gray!20}{Update}: All task execution information---including task specifications, coordinator action messages, and sub-agent inputs and outputs---is sequentially pushed onto the stack.
    \item \colorbox{gray!20}{Condensation}: When the coordinator determines that the memory becomes verbose or that a task stage has been completed, \textsc{Revise} performs \emph{memory condensation} by popping a contiguous segment $\{m_i\}_{i=k}^t$ from the stack, summarizing it into a compact representation $m'$, and pushing $m'$ back onto the stack. This operation preserves task information while reducing redundancy.
    \item \colorbox{gray!20}{Pruning}: When the coordinator detects unproductive or erroneous exploration, \textsc{Revise} performs \emph{memory pruning} by removing a selected segment of memory entries from the stack. Additionally, a concise record of failure causes is retained to guide subsequent exploration.
\end{itemize}

By exposing memory as an explicit control target, \textsc{Revise} 
enables active memory optimization, effectively 
filtering noise and correcting earlier coordination errors with minimal overhead. Implementation details of \textsc{Revise} are in Appendix~\ref{appendix: prompts}.

\subsection{Structured Experience Memory}

To support cross-task generalization, we maintain a structured 
\textbf{experience memory} that stores persistent information beyond individual 
task executions. The experience memory consists of three complementary components: 
(i) \emph{user profiles}, which capture stable user attributes and preference 
signals; (ii) \emph{semantic memory}, which stores factual knowledge and 
declarative information, particularly externally retrieved evidence; and 
(iii) \emph{procedural memory (SOPs)}, which abstracts key execution steps from 
previously completed tasks as reusable procedural patterns. These components are 
organized with a unified storage and retrieval interface.
Examples of experience memory entries, storage formats and prompting details, are in Appendix~\ref{appendix: prompts}.

\paragraph{Experience Retrieval}
We further design an Experience Search agent queries the experience memory using the current task 
representation and user identifier, retrieving relevant entries that are summarized and injected into the task memory to inform coordination and mitigate cold-start issues.

\paragraph{Reinforcement Learning Formulation}
\label{sec:rl}
We formulate training \mname's coordinator as a multi-step RL problem,
where the policy model is augmented with access to an external search engine and a structured memory stack.
Given a query $q \sim \mathcal{D}$,
the policy model $\pi_\theta$ generates a trajectory
$
y = (a_1, \dots, a_T)
$ with $T$ action steps, and the RL objective with search engine invocations and memory stack operations is defined as:
\begin{equation}
\small
\begin{aligned}
\text{ }&\max_{\theta}\;\;
\mathbb{E}_{q \sim \mathcal{D},\, y \sim \pi_\theta(\cdot \mid q;\mathcal{R},\mathcal{M})}
\!\left[ r_\phi(q,y) \right] \\
&\;-\; \beta\,
\mathbb{D}_{\mathrm{KL}}\!\bigl(
\pi_\theta(y\mid q;\mathcal{R},\mathcal{M})
\,\|\,
\pi_{\mathrm{ref}}(y \mid q;\mathcal{R},\mathcal{M})
\bigl),
\end{aligned} 
\label{eq:rl_objective}
\end{equation}
where $\mathcal{R}$ and $\mathcal{M}$ denotes search engine and stack-structured memory respectively,
$r_\phi$ is the reward function,
and $\pi_{\mathrm{ref}}$ is the frozen reference policy.
Here, both $\mathcal{R}$ and $\mathcal{M}$ are dynamically updated along the trajectory, and the policy is conditioned on the interaction state $s_i = (\mathcal{R}_i, \mathcal{M}_i)$ at each step, which captures the current search results and memory stack.
Unlike standard RLHF~\cite{schulman2017proximal} or retrieval-augmented RL methods such as Search-R1~\cite{jin2025search},
which largely rely on parametric knowledge and coarse-grained searching interactions, our policy follows an interleaved \emph{\textbf{retrieval--reasoning--memory}} execution paradigm.
Concretely, $\pi_\theta(\cdot \mid q;\mathcal{R}, \mathcal{M})$ can be viewed as a sequence of $T$ alternating reasoning, searching and memorizing actions,
where each step conditions only on information obtained through retrieval
or reasoned and kept in the memory stack.

We adopt
\textbf{Group Relative Policy Optimization (GRPO)}~\cite{shao2024deepseekmath}
to optimize the policy, which eliminates the need for a learned value function by computing
relative advantages from statistics of the current rollout group.
Specifically, for a rollout group consisting of $K$ trajectories
$\{y^{(k)}\}_{k=1}^{K}$ sampled from the old policy $\pi_{\theta_{\mathrm{old}}}$,
where each trajectory
$
y^{(k)} = (x^{(k)}_1, \dots, x^{(k)}_{|y^{(k)}|})
$
is a sequence of generated tokens\footnote{In our implementation, each high-level action $a_t$ is realized as a contiguous sequence of generated tokens},
let $\mathcal{R}_G$ denote the set of all token-level rewards
$\{ r^{(k)}_i\}$.
In particular, we define token-level rewards as
\begin{equation}
\small
r^{(k)}_i = r^{(k)}_{\text{out}} + I^{(k)}_{\text{act},i} \cdot r^{(k)}_{\text{mem},i},
\end{equation}
where $r^{(k)}_{\text{out}}$ is the trajectory-level result reward (e.g., EM/F1 accuracy), $r^{(k)}_{\text{mem},i}$ is the process reward associated with \textsc{Revise} operations, and $I^{(k)}_{\text{act},i}$ is an indicator function that equals 1 if token $x^{(k)}_i$ belongs to a \textsc{Revise} action span, and 0 otherwise. This provides a deterministic mapping from sparse action-level rewards to token-level supervision.

For each token $x^{(k)}_i$ in trajectory $y^{(k)}$,
we compute a normalized group-relative advantage as:
\begin{equation}
\footnotesize
\hat{A}^{(k)}_i
=
\Bigl({
r^{(k)}_i - \texttt{mean}(\mathcal{R}_G) \Bigl) /
}{
\texttt{std}(\mathcal{R}_G)
}.
\end{equation}
This group-relative normalization avoids explicit value-function learning and provides stable advantage estimation in the presence of dynamically evolving search and memory states.

\begin{table*}[t]
\small
\setlength{\tabcolsep}{6pt}
\renewcommand{\arraystretch}{1.1}
\centering
\begin{tabular}{l l | c c c c | c c c c}
\toprule
\rowcolor{gray!30}
\multicolumn{2}{c|}{\textbf{Method}} &
\multicolumn{4}{c|}{\textbf{Qwen2.5-3B}} &
\multicolumn{4}{c}{\textbf{Qwen2.5-7B}} \\

\rowcolor{gray!30}
\textbf{Paradigm} & \textbf{Approach} &
\textbf{2Wiki} & \textbf{MusiQue} & \textbf{GAIA} & \textbf{FRAMES} &
\textbf{2Wiki} & \textbf{MusiQue} & \textbf{GAIA} & \textbf{FRAMES} \\
\midrule

% ================= Naive =================
\multirow{2}{*}{Naive}
& Base
& 23.98 & 9.70 & 5.70 & 8.01
& 25.41 & 12.15 & 4.29 & 12.52 \\

& FS-RAG
& 15.47 & 7.64 & 4.30 & 10.42
& 17.71 & 10.74 & 5.02 & 12.52 \\
\midrule

% ================= Single-Agent =================
\multirow{2}{*}{Single-Agent}
& ReACT
& 25.09 & 13.92 & 4.78 & 10.53
& 27.51 & 19.34 & 6.37 & 15.29 \\

& IRCoT
& 15.89 & 12.43 & 2.77 & 6.79
& 36.45 & 8.39 & 5.50 & 6.78 \\
\midrule

% ================= Multi-Agent =================
\multirow{3}{*}{Multi-Agent}

& OWL
& 17.39 & 14.81 & 3.28 & 13.49
& 29.73 & 17.66 & 5.39 & 14.68 \\

& MacNet
& 25.20 & 13.19 & / & 11.92
& 28.19 & 17.81 & / & 12.61 \\

& AFlow
& 24.56 & 13.07 & 2.57 & 12.13
& 30.53 & 18.15 & 4.72 & 12.81 \\
\midrule

% ================= Agentic-RL =================
\multirow{3}{*}{Agentic-RL}
& ReSearch
& 27.23 & 9.47 & 4.48 & 10.00
& 30.03 & 12.58 & 4.43 & 15.61 \\

& ARPO
& 29.55 & 13.38 & \textbf{7.71} & 13.49
& 30.71 & 12.71 & 8.56 & 12.18 \\

& \textbf{\mname{}}
& \textbf{32.92} & \textbf{16.48} & \textbf{7.71} & \textbf{16.23}
& \textbf{38.34} & \textbf{22.01} & \textbf{9.45} & \textbf{19.44} \\

\bottomrule
\end{tabular}

% \caption{Performance comparison on multi-hop QA benchmarks across different paradigms and model scales.}
% \caption{
% Performance comparison (\textbf{F1}, \%) on multi-hop QA benchmarks
% (\textit{2WikiMultiHopQA}, \textit{MusiQue}, \textit{GAIA}, and \textit{FRAMES})
% across different paradigms using \textbf{Qwen2.5-3B} and \textbf{Qwen2.5-7B},
% where \textbf{bold} indicates the best result.
% }
\vspace{-0.2cm}
\caption{
Performance comparison (\textbf{F1}, \%) on multi-hop QA benchmarks
(\textit{2WikiMultiHopQA}, \textit{MusiQue}, \textit{GAIA}, and \textit{FRAMES})
across different paradigms using \textbf{Qwen2.5-3B} and \textbf{Qwen2.5-7B}.
% The symbol ``/'' indicates that a model could not produce results on a dataset,
% and \textbf{bold} highlights the best performance in each column.
}
\label{tab:comparison}
\vspace{-0.4cm}
\end{table*}

GRPO optimization objective is then defined as:
\begin{equation}
\small
\begin{split}
\mathcal{J}(\theta)
=
\mathbb{E}\Big[
\frac{1}{K} \sum_{k=1}^{K}
\frac{1}{|y^{(k)}|}
\sum_{i=1}^{|y^{(k)}|}
\texttt{Clip}(\tilde{z}_{i}^{(k)}, \hat{A}^{(k)}_i )
\Big]
-
\beta \mathbb{D}_{\mathrm{KL}}, 
\end{split}
\notag
\end{equation}
$\texttt{Clip}(\tilde{z}_{i}^{(k)}, \hat{A}^{(k)}_i)
=
\min (
\tilde{z}_{i}^{(k)} \hat{A}^{(k)}_i,
\operatorname{clip}(\tilde{z}_{i}^{(k)}, 1\pm\varepsilon)\hat{A}^{(k)}_i
)$, importance ratio \(
\tilde{z}_{i}^{(k)}=\frac{
\pi_\theta(x^{(k)}_i \mid q, x^{(k)}_{<i}; \mathcal{R}, \mathcal{M})
}{
\pi_{\theta_{\mathrm{old}}}(x^{(k)}_i \mid q, x^{(k)}_{<i}; \mathcal{R}, \mathcal{M})
}\) denotes the probability ratio at the token level.
Term $\mathbb{D}_{\mathrm{KL}}(\pi_\theta \| \pi_{\mathrm{ref}})$
constrains the updated policy to remain close to a frozen reference policy
$\pi_{\mathrm{ref}}$.
Notably, although $\textsc{Revise}$ introduces additional process-level rewards, the dominant optimization signal remains the trajectory-level result reward, ensuring that all memory operations are ultimately aligned with final answer correctness.
Notably, all rewards, advantages, and policy updates are defined at action level and applied at \emph{token level}.

\paragraph{Human-in-the-Loop Coordination}
We further model human interaction as a privileged memory update in the coordinator's decision process. Let the coordinator state be $s_t=(\mathcal{M}_t,\mathcal{E}_t,g_t)$, where $\mathcal{M}_t$ is the task-memory stack, $\mathcal{E}_t$ is retrieved experience memory, and $g_t$ is the current goal. Human feedback $h_t$ can be introduced either by SOP-triggered checkpoints retrieved from experience memory, or by the coordinator's own uncertainty-aware delegation:
\[
a_t=\textsc{Delegate}(\textsc{Human}, r_t),
\]
where $r_t$ specifies the required clarification, confirmation, or preference signal. The returned feedback is normalized and pushed into task memory,
\[
\mathcal{M}_{t+1}=\textsc{Push}(\mathcal{M}_t,\phi_h(h_t)).
\]
Different from ordinary sub-agent observations, human feedback is treated as a high-priority constraint for subsequent decisions:
\[
\pi_\theta(a_{t+1}\mid \mathcal{M}_{t+1},\mathcal{E}_t,g_t)
\quad \text{s.t.} \quad a_{t+1}\models h_t .
\]
This mechanism also supports post-completion refinement. Given a completed trajectory with final memory $\mathcal{M}_T$, new user feedback $h^{\mathrm{fb}}$ reopens the coordination loop as
\[
\mathcal{M}'_0=\textsc{Push}(\mathcal{M}_T,\phi_h(h^{\mathrm{fb}})),
\]
after which the coordinator continues planning from the preserved task state and invokes sub-agents only for the parts affected by the feedback. Thus, SOP-triggered questioning, coordinator-initiated clarification, and multi-turn user revision are handled by the same memory coordination.

\begin{table*}[t]
\small
\setlength{\tabcolsep}{4pt}
\renewcommand{\arraystretch}{1.1}
\centering
\resizebox{1\textwidth}{!}{%
\begin{tabular}{l | c c c c c | c c | c c c c c | c c}
\toprule
\rowcolor{gray!30}
\textbf{Approach} &
\multicolumn{7}{c|}{\textbf{Qwen3-32b}} &
\multicolumn{7}{c}{\textbf{DeepSeek-V3.2}} \\

\rowcolor{gray!30}
& \multicolumn{5}{c}{\textbf{RACE}} &
\multicolumn{2}{c|}{\textbf{FACT}} &
\multicolumn{5}{c}{\textbf{RACE}} &
\multicolumn{2}{c}{\textbf{FACT}} \\

\rowcolor{gray!30}
& \textbf{Over.} & \textbf{Comp} & \textbf{Ins.} & \textbf{Inst} & \textbf{Read} &
\textbf{Eff.c.} & \textbf{C.acc.} &
\textbf{Over.} & \textbf{Comp} & \textbf{Ins.} & \textbf{Inst} & \textbf{Read} &
\textbf{Eff.c.} & \textbf{C.acc.} \\
\midrule

Prompt    & 39.54  & 36.87  & 35.25  & 45.34 &  43.67  & / & /
% & - & - & - & - & - & -- & -- \\
& 41.81 & 39.81 & 41.31 & 44.92 & 42.25  & / & / \\
RAG       & 37.94 & 34.79 & 33.76 & 43.27 & 43.08& 3.46  & 12.73
% & - & - & - & - & - & - & -  \\
& 41.10 & 40.14 & 39.83 & 42.71 & 39.10 & 2.20  & 11.00 \\
\midrule
ReAct     & 41.14 & 39.40 & 37.23 & 45.19 & 43.81 & 4.76 & 24.09
% & - & - & - & - & - & - & - \\
& 43.30 & 42.86 & 42.26 & 44.81 & 43.71 & 7.17  & 15.32  \\
IRCoT     & 41.47 & 39.52 & 38.38 & 45.32 &45.05 & 11.36 & 28.21 
% & - & - & - & - & - & - & - \\
& 41.57 & 38.60 & 39.02 & 46.21 & 42.71 & 31.50 & 74.72 \\
\midrule

STORM       & 35.57 & 33.79 & 31.91 & 41.12 & 37.23 & 8.67 & 26.90 
% & - & - & - & - & - & - & - \\
& 43.69 & 42.30 & 41.85 & 46.09 & 46.12 & 2.64  & 32.58 \\
WebWeaver &  40.52 & 38.81 & 38.28 & 45.34 & 41.07 & 8.04 &25.42 
& 43.25& 40.94 & 41.98 & 47.07 & 44.79& 29.63 & \textbf{84.52}\\
EDR       & 40.40 & 38.44 & 34.15 & 45.18 & 37.08 & 3.14 & 12.08 & 42.28 & 41.02 & 39.28 & 45.36 & 44.59 & 16.53 & 27.56 \\

\midrule

\textbf{\mname{}} & \textbf{42.55} & \textbf{40.85} & \textbf{38.92} & \textbf{47.27} & \textbf{45.50}  & \textbf{14.82} & \textbf{29.33} & \textbf{47.47} & \textbf{47.18} & \textbf{46.91} & \textbf{48.93 } & \textbf{47.29} & \textbf{32.75} &50.50 \\
\bottomrule
\end{tabular}}
\caption{
Report generation performance on DeepResearch Bench using \textbf{Qwen3-32B} and \textbf{DeepSeek-V3.2} as backbone LLMs.
RACE evaluates report quality; FACT measures factual grounding via effective citation rate (Eff.c.) and citation accuracy (C.acc.).
``/'' indicates the metric is not applicable.
}
\label{tab:report_generation}
\vspace{-0.3cm}
\end{table*}

 \begin{table}[t]
 \centering
 \fontsize{7.5pt}{9.5pt}\selectfont
 \setlength{\tabcolsep}{2.4pt}
 \renewcommand{\arraystretch}{1.1}
 \resizebox{0.49\textwidth}{!}{%
 \begin{tabular}{l|ccccc|c}
 \toprule
 % \rowcolor{gray!10}
 \rowcolor{gray!30}
 \textbf{Setting} & \textbf{Ins.}$\uparrow$ & \textbf{Evi.}$\uparrow$ & \textbf{Log.}$\uparrow$ & \textbf{Com.}$\uparrow$ & \textbf{Exp.}$\uparrow$ & \textbf{Tot.}$\uparrow$ \\
 \midrule
 ReAct-FW      & 14.03 & 11.98 & 13.22 &  5.24 & 11.72 & 56.19 \\
 ReAct-GR      & 12.42 & 11.25 & 10.85 & 11.50 & 13.61 & 59.63 \\
 \midrule
 \textbf{\mname{}}   & \textbf{14.35} & \textbf{13.26} & \textbf{15.55} & \textbf{14.42} & \textbf{13.53} & \textbf{71.11} \\
 \bottomrule
 \end{tabular}}
 \caption{Multi-turn refinement on 100 follow-up instances from DeepResearch Bench. Each dimension is scored 0--20; Tot.\ is their sum.}
 \label{tab:multiturn}
 \end{table}

\begin{table}[htbp]
\centering
\footnotesize
\setlength{\tabcolsep}{6pt}
\renewcommand{\arraystretch}{1.1}
\resizebox{0.49\textwidth}{!}
{\begin{tabular}{l | c c c c}
\toprule
\rowcolor{gray!30}
\textbf{Method} &
\textbf{2Wiki} & \textbf{Musique} &
\textbf{GAIA} & \textbf{FRAMES} \\

\midrule

% ================= Full Model =================
\textbf{\mname{}} & \textbf{32.92} & \textbf{16.48} & \textbf{7.71} & \textbf{16.23} \\

\midrule

% ================= Component Ablation =================
w/o Task Memory      & 29.90 & 10.76 & 4.68 & 13.53 \\
w/o Experience       & 28.47 & 8.99  & 5.53 & 7.69 \\
w/o Both          & 17.12 & 7.43  & 2.47 & 6.33 \\

\bottomrule
\end{tabular}}

\caption{Ablation analysis of component and reward designs in \mname~ on Qwen2.5-3B.}
\label{tab:ablation}
\vspace{-0.5cm}
\end{table}

% \begin{table*}[t]
% \centering
% \small
% \setlength{\tabcolsep}{4pt}
% \renewcommand{\arraystretch}{1.1}
% \begin{tabular}{lcccccc}
% \toprule
% \rowcolor{gray!10}
% \textbf{Method} & \textbf{GSM8K} & \textbf{Math} & \textbf{TriviaQA} & \textbf{PopQA} & \textbf{GPQA} & \textbf{Avg.} \\
% \midrule
% StackPlanner-rl-sop & 57.6 & 42.3 & 76.3 & 42.3 & 11.8 & 46.06 \\
% StackPlanner-rl & 56.0 & 45.8 & 76.9 & 44.7 & 17.6 & 48.20 \\
% \textbf{Full} & \textbf{61.2} & \textbf{50.7} & \textbf{77.7} & \textbf{49.0} & \textbf{23.8} & \textbf{52.48} \\
% \bottomrule
% \end{tabular}
% \caption{Experience-retrieval ablation across QA and reasoning datasets.}
% \label{tab:retrieval_ablation}
% \end{table*}

% \begin{table}[t]
% \centering
% \footnotesize
% \setlength{\tabcolsep}{6pt}
% \renewcommand{\arraystretch}{1.1}
% \resizebox{0.49\textwidth}{!}
% {\begin{tabular}{l | c c c c}
% \toprule
% \rowcolor{gray!30}
% \textbf{Method} & \textbf{TriviaQA} & \textbf{PopQA} & \textbf{GPQA} & \textbf{Avg.} \\
% \midrule
% \textbf{Ours} & \textbf{77.7} & \textbf{49.0} & \textbf{23.8} & \textbf{50.17} \\
% \midrule
% w/o RL& 76.9 & 44.7 & 17.6 & 46.40 \\
% w/o RL\&SOP  & 76.3 & 42.3 & 11.8 & 43.47 \\
% \bottomrule
% \end{tabular}}
% \caption{Experience-retrieval ablation on knowledge-intensive QA and reasoning datasets.}
% \label{tab:retrieval_ablation}
% \end{table}

\begin{table}[t]
\centering
\small
\setlength{\tabcolsep}{6pt}
\renewcommand{\arraystretch}{1.15}
\label{tab:retrieval_ablation}
\vspace{-0.15cm}
\resizebox{0.49\textwidth}{!}{
\begin{tabular}{lccccc}
\rowcolor{gray!30}
\toprule 
Method
& GSM8K
& MATH
& TriviaQA
& PopQA
& GPQA \\

\midrule

% \rowcolor{gray!8}
\mname{}
& \textbf{61.2}
& \textbf{50.7}
& \textbf{77.7}
& \textbf{49.0}
& \textbf{23.8} \\

w/o RL
& 56.0
& 45.8
& 76.9
& 44.7
& 17.6\\

w/o RL + Experience
& 57.6
& 42.3
& 76.3
& 42.3
& 11.8\\

\bottomrule
\end{tabular}}
\caption{
Experience-retrieval ablation on
knowledge-intensive QA and reasoning benchmarks.
}
\vspace{-0.35cm}
\end{table}

\begin{table*}[t]
\centering
\footnotesize
\setlength{\tabcolsep}{6pt}
\renewcommand{\arraystretch}{1.12}
\resizebox{0.7\textwidth}{!}{%
\begin{tabular}{l | c c c c c c c | c c}
\toprule
\rowcolor{gray!30}
\textbf{Method} & \multicolumn{7}{c|}{\textbf{ALFWorld}} & \multicolumn{2}{c}{\textbf{WebShop}} \\
% \cmidrule(lr){2-8} \cmidrule(lr){9-10}
\rowcolor{gray!30}
& \textbf{Pick} & \textbf{Look} & \textbf{Clean} & \textbf{Heat} & \textbf{Cool} & \textbf{Pick2} & \textbf{Avg.} & \textbf{Score} & \textbf{Acc.} \\
\midrule
Prompt & 45.8 & 48.4 & 34.5 & 41.0 & 19.6 & 31.7 & 36.5 & 14.7 & 0.8 \\
\textbf{\mname{}} & \textbf{84.7} & \textbf{80.6} & 24.1 & 33.3 & \textbf{37.0} & \textbf{51.2} & \textbf{51.1} & \textbf{27.8} & \textbf{2.3} \\
\bottomrule
\end{tabular}}
\caption{Generalization of \mname{} to interactive decision-making environments.}
\label{tab:interactive_generalization}
\vspace{-0.5cm}
\end{table*}

   \vspace{-0.15cm}
  \section{Experiment}
  \vspace{-0.15cm}
                                                                                                                                                        
  We evaluate \mname{} on both answer-oriented deep-search benchmarks and                                                                             
  open-ended deep-research tasks to verify whether explicit task-memory control                                                                         
  and reusable experience memory improve long-horizon coordination,
  generalization, and information organization:

  \begin{itemize}[leftmargin=*,noitemsep,topsep=2pt]
      \item \textbf{RQ1:} Does \mname{} outperform representative baselines on
            multi-hop QA and agentic benchmarks?
      \item \textbf{RQ2:} Does \mname{} produce higher-quality and
            better-grounded reports on open-ended deep-research tasks?
      \item \textbf{RQ3:} Can \mname{} support human-in-the-loop refinement by
            incorporating multi-turn follow-up feedback into an existing task
            state?
      \item \textbf{RQ4:} How do task and experience memory each
            contribute to final performance?
  \end{itemize}

  \subsection{Experimental Setup}
  \textbf{\ding{182} Evaluation Benchmarks.}
  We evaluate \mname{} across two answer-oriented settings:
  \emph{multi-hop QA} (\textit{2WikiMultiHopQA}~\cite{ho2020constructing} and
  \textit{MuSiQue}~\cite{trivedi2022musique}) and \emph{agentic benchmarks}
  (\textit{GAIA}~\cite{mialon2023gaia} and
  \textit{FRAMES}~\cite{krishna2024retrieval}).
  For open-ended deep research, we additionally use
  \textbf{DeepResearch Bench}~\cite{du2025deepresearch} to assess long-form
  report generation.
  Further dataset details are provided in Appendix~\ref{appendix:datasets}.
  \textbf{\ding{183} Baselines.}
  We compare \mname{} against a diverse set of baselines spanning four
  paradigms.
  \emph{Naive} baselines include Base and
  FS-RAG~\cite{trivedi2023interleaving}.
  \emph{Single-agent} approaches consist of ReAct~\cite{react} and
  IRCoT~\cite{trivedi2023interleaving}.
  \emph{Multi-agent} methods cover both centralized architectures
  (OWL~\cite{hu2025owl}) and automated pipeline architectures
  (MacNet~\cite{qianscaling} and AFlow~\cite{zhangaflow}).
  \emph{Agentic-RL} baselines include ReSearch~\cite{chen2025learning} and
  ARPO~\cite{dong2025agentic2}.
  For report generation, we additionally compare against
  STORM~\cite{shao2024assisting}, WebWeaver~\cite{li2025webweaver}, and
  EDR~\cite{prabhakar2025enterprise}.
  Detailed descriptions of all baselines are provided in
  Appendix~\ref{appendix:baselines}.
  \textbf{\ding{184} RAG Tools.}
  We use a Wikipedia-based search tool (snapshot: November 1, 2023) and Bocha
  for web search.

  \subsection{Main Result Analysis}

  To answer \textbf{RQ1}, we evaluate whether \mname{} improves performance on
  answer-oriented deep-search tasks.
  As shown in Table~\ref{tab:comparison}, \mname{} achieves state-of-the-art
  results across all benchmarks, outperforming baselines on both multi-hop QA
  and agentic evaluation.
  It demonstrates strong out-of-distribution generalization on \textit{MuSiQue},
  \textit{GAIA}, and \textit{FRAMES}, achieving F1 scores of 16.48\%, 7.71\%,
  and 16.23\% with Qwen2.5-3B, and 22.01\%, 9.45\%, and 19.44\% with
  Qwen2.5-7B, respectively.
  \textit{GAIA} is the most challenging benchmark owing to its demands for
  multi-step reasoning and long-horizon memory management: baselines such as
  MacNet fail to produce results (marked ``/'') because they lack effective
  task-memory mechanisms, while AFlow achieves only 2.57\% and 4.72\% on the
  two backbones.
  By contrast, \mname{} handles complex reasoning and memory management
  effectively, delivering consistently strong results.

  \subsection{Open-Ended Deep Research}

  To answer \textbf{RQ2}, we evaluate \mname{} on \textbf{DeepResearch Bench},
  which measures report quality via the RACE metric (Overall,
  Comprehensiveness, Insight, Instruction-Following, Readability) and factual
  grounding via FACT (Effective Citation Rate and Citation Accuracy).
  For a fair comparison, every column in Table~\ref{tab:report_generation} uses
  the same report-generation backbone across all methods; in \mname{}, this
  backbone is used by the sub-agents, while the coordinator remains the
  RL-trained Qwen2.5-7B policy and retrieves procedural experience (SOPs) to
  guide planning.
  \mname{} achieves the strongest overall report quality under both Qwen3-32B
  and DeepSeek-V3.2: it attains the best score on every RACE dimension with
  Qwen3-32B (42.55 Overall, 14.82 Effective Citation Rate, 29.33 Citation
  Accuracy), and further improves to 47.47 Overall and 32.75 Effective Citation
  Rate under DeepSeek-V3.2.
  Although WebWeaver yields higher Citation Accuracy under DeepSeek-V3.2,
  \mname{} strikes a better balance between report quality and citation
  effectiveness, suggesting that a fixed, RL-trained coordinator can leverage
  retrieved procedural guidance to orchestrate stronger generation backbones
  more effectively.

  \subsection{Multi-Turn Interaction}

  To answer \textbf{RQ3}, Table~\ref{tab:multiturn} evaluates whether
  \mname{} can handle user follow-up requests by incrementally updating an
  existing deep-research report.
  We compare against two ReAct-based baselines: \textit{ReAct-FW} writes a
  fresh report from a query that concatenates the original and follow-up
  instructions, while \textit{ReAct-GR} conditions on the first-turn report and
  globally rewrites it in accordance with the follow-up.
  ReAct-FW remains instruction-following but discards substantial prior content,
  resulting in a low completeness score (Com.\ 5.24).
  ReAct-GR retains more prior content (Com.\ 11.50), yet global rewriting
  compromises logical consistency (Log.\ 10.85).
  In contrast, \mname{} reuses the existing task memory and applies targeted
  revision only to the portions of the report affected by the follow-up,
  achieving highest total score (71.11) as well as the best scores on
  instruction following, evidence grounding, logical consistency, and
  completeness.
  Results indicate that task-memory updates provide a principled
  mechanism for non-destructive multi-turn refinement.
    \vspace{-0.5cm}
  \subsection{Memory Ablation Study}

  \textbf{\ding{182} Memory-Component Ablation.}
  To answer \textbf{RQ4} (component contribution), we ablate the task memory and
  experience memory modules while keeping the remaining architecture and training
  setup unchanged.
  As reported in Table~\ref{tab:ablation}, removing task memory leads to
  performance drops of 3.02\%, 5.72\%, 3.03\%, and 2.70\% on
  \textit{2WikiMultiHopQA}, \textit{MuSiQue}, \textit{GAIA}, and
  \textit{FRAMES}, respectively, indicating that the coordinator requires an
  actively maintained task state to preserve intermediate reasoning progress
  rather than relying solely on a raw interaction trace.
  Removing experience memory causes larger declines of 4.45\%, 7.49\%, 2.18\%,
  and 8.54\% on the same datasets, demonstrating that reusable SOPs are
  important for planning, retrieval decisions, and cross-task generalization.
  Removing both components yields the largest degradation across all benchmarks,
  with F1 scores falling by 15.80\%, 9.05\%, 5.24\%, and 9.90\%, respectively.
  Together, these results confirm that task memory and experience memory play
  complementary roles: task memory maintains current execution state, while
  experience memory transfers reusable procedures from prior tasks.

    \paragraph{\ding{183} Experience-Retrieval Ablation.}                                                                                                 
  To answer \textbf{RQ4} (retrieval contribution), Table 5
  compares three variants that differ in coordinator training and experience memory on GSM8K~\cite{gsm8k}, MATH~\cite{math}, TriviaQA~\cite{TriviaQA}, PopQA~\cite{PopQA}, GPQA~\cite{GPQA} datasets.                                                                 
  \textit{w/o RL + Experience} removes both the RL-trained coordinator and                                                                            
  experience memory retrieval, relying solely on fixed SOPs, and achieves an
  average score of 46.06.
  \textit{w/o RL} re-introduces experience memory retrieval while keeping the
  coordinator untuned, improving the average to 48.20 and confirming that
  dynamically retrieved experience provides additional cross-task guidance
  beyond static SOPs.
  The full \mname{}, which combines the RL-trained coordinator with the
  two-channel MemRetriever, further raises the average to 52.48.
  The gain is most pronounced on \textit{GPQA}, where performance increases
  from 11.8 to 17.6 and then to 23.8, suggesting that knowledge-intensive
  reasoning tasks benefit most from accurate retrieval of reusable procedural
  experience.
  These results show that experience memory is not a static knowledge bank;
  the full gains require both the retrieval mechanism and an RL-trained
  coordinator.
  
  \paragraph{\ding{184} Generalization to Interactive Tasks.}
  We further examine whether the memory-control mechanism of \mname{} generalizes
  beyond text-only QA.
  Table~\ref{tab:interactive_generalization} compares Qwen3-32B with and without
  \mname{} on ALFWorld and WebShop, two interactive decision-making environments
  that require maintaining goals, observations, and action history across
  multiple steps.
  Incorporating \mname{} improves ALFWorld Avg.\ from 36.5 to 51.1 and WebShop
  Score from 14.7 to 27.8, with Acc.\ increasing from 0.8 to 2.3.
  Gains are especially pronounced on ALFWorld Pick, Look, Cool, and Pick2
  tasks, where persistent state tracking is critical.
  The consistent improvements across both environments suggest that explicit
  task-state and experience-memory control is broadly beneficial, extending well
  beyond text-based reasoning to more general agentic settings.

% \section{Conclusion}
\vspace{-0.2cm}
\section{Conclusion and Future Work}
We present \mname, a hierarchical centralized multi-agent framework that treats memory as an explicit
  control target for coordination. By combining decoupled coordination with active \textbf{task memory}
  management and reusable \textbf{experience memory}, \mname\ mitigates context bloat and error propagation
   in long-horizon collaboration. Moreover, high-level coordination and memory control are jointly
  optimized via RL.
  Several directions remain for future work. First, designing more expressive yet compact task memory may further improve decision robustness under longer horizons and more complex inter-agent
  interactions. Second, exploring latent representations for inter-agent communication could reduce
  coordination overhead and accelerate multi-agent inference. 
  % Finally, enabling multi-agent systems to
  % autonomously evolve their strategies in complex, non-stationary environments, represents a promising step toward truly adaptive and
  % self-improving agent collectives.
\newpage

\section*{Limitations}
Despite the promising results, our framework does have some limitations that need to be addressed.
% \ding{182} \textbf{Limited support for multi-turn interactions.} The current task-level memory is primarily designed for single-turn and does not explicitly model multi-turn conversational dependencies. As a result, adapting the behavior of specific sub-agents across extended interactions becomes cumbersome and error-prone.
 \ding{182} \textbf{Limited coverage of extended human-in-the-loop interaction.} Although \mname{} supports follow-up refinement by incrementally updating an existing task state, our current evaluation focuses on short follow-up scenarios with localized user intents. More complex human-in-the-loop settings, such as long interaction chains, ambiguous preference updates, or repeated shifts in user goals, remain underexplored. Extending the task memory to robustly track such evolving conversational dependencies is an important direction for future work.
\ding{183} \textbf{Cold-start challenges in experience memory.} Experience memory mechanisms still suffer from cold-start issues, where insufficient prior experience limits their effectiveness in early stages. While simulated users can be introduced to partially mitigate this problem, the initialized experiences often exhibit limited generalization capability when transferred to real or diverse user behaviors.

\section*{Ethical considerations}

All experiments in this study were conducted solely on publicly available benchmark datasets, including \textit{2WikiMultiHopQA}, \textit{MuSiQue}, \textit{GAIA}, and \textit{FRAMES}, in compliance with their respective licenses and usage terms. We did not utilize any personally identifiable information, nor were any human or animal subjects involved in the research.

% \bibliography{custom}
\bibliography{custom}

\newpage
\appendix
\newpage

\section{Experiment Datasets}
\label{appendix:datasets}
% \subsection{Dataset Construction and Sampling Strategies}
\subsection{Training Dataset.}
Followed by ~\cite{gao2025turnsunlockinglonghorizonagentic}, We train our models and baselines on a curated multi-hop question answering dataset
constructed from the training splits of 2WikiMultiHopQA~\cite{ho2020constructing}.
To focus on genuinely non-trivial reasoning scenarios,
we filter out instances that require no external retrieval
or can be solved with only a single, trivial retrieval step.

\subsection{Testing set.} 
\begin{table}[ht]
\centering
\small
\setlength{\tabcolsep}{6pt}
\renewcommand{\arraystretch}{1.1}
\begin{tabular}{l|c|c|c}
\toprule
\rowcolor{gray!10}
\textbf{Dataset} & \textbf{Train} & \textbf{Dev} & \textbf{Test} \\
\midrule
2Wiki& 154,878 & 12,576 & 12,576 \\
MuSiQue  & 19,938 & 2,417 & 2,459 \\
GAIA  & 0 & 0 & 127 \\
FRAMES & 0 & 0 & 824 \\
\bottomrule
\end{tabular}
\caption{Overview of datasets used in experiments.}
\label{tab:dataset_overview}
\end{table}

We evaluate our approach on four widely used benchmarks covering multi-hop QA, and real-world agent evaluation:
Key statistics for training, development, and test splits are summarized in Table~\ref{tab:dataset_overview}.

\paragraph{\ding{182} Multi-Hop QA Benchmarks.} 
We evaluate our approach on two multi-hop question answering datasets that require reasoning over multiple documents: 
\begin{itemize}[leftmargin=*]
    \item \textbf{2WikiMultiHopQA~\cite{ho2020constructing}.} 
    Constructed from Wikipedia and Wikidata, this dataset contains 192,606 question--answer pairs. 
    It includes 154,878 training, 12,576 development, and 12,576 test instances, 
    focusing on tasks that necessitate aggregating evidence across multiple sources.
    \item \textbf{MuSiQue~\cite{trivedi2022musique}.} 
    MuSiQue is designed to test multi-step reasoning over Wikipedia data, 
    with each reasoning step depending critically on the previous step. 
    The dataset comprises 19,938 training, 2,417 development, and 2,459 test examples.
\end{itemize}

\paragraph{\ding{183} Agentic Benchmarks.} 
We further assess our method on two agentic benchmarks that evaluate models’ ability to handle real-world questions:
\begin{itemize}[leftmargin=*]
    \item \textbf{GAIA~\cite{mialon2023gaia}.} 
    GAIA measures performance on tasks requiring multi-step reasoning, web interaction, and multi-modal input handling. 
    We choose 127 text-only questions in validation set across varying difficulty levels.
    \item \textbf{FRAMES~\cite{krishna2024retrieval}.} 
    FRAMES consists of 824 multi-hop questions, 
    emphasizing factual accuracy, retrieval, and reasoning over multiple sources.
\end{itemize}

% \end{itemize}

\begin{table}[h]
\centering
\resizebox{0.49\textwidth}{!}{\begin{tabular}{lc}
\toprule
Method & Resolved (\%) \\
\midrule
mini-SWE-agent +Qwen2.5-Coder 32B & 9.00 \\
StackPlanner + Qwen2.5 7B & 9.12 \\
\bottomrule
\end{tabular}}
\caption{Performance on SWE-bench-verified.}
\label{table:swe-bench}
\end{table}

\paragraph{\ding{184} Deep Research Benchmark.}
 Beyond answer-oriented evaluation, we test \mname{} on \textbf{DeepResearch Bench}~\cite{du2025deepresearch}, which is designed to evaluate agents in open-ended research report generation. The benchmark includes 100 expert-level tasks, evenly split between English and Chinese, and covers 22 domains selected to reflect the distribution of large-scale real user research requests. Its evaluation protocol considers both writing quality and factual grounding: \textbf{RACE} measures the overall quality of generated reports from aspects such as coverage, insightfulness, instruction adherence, and readability, while \textbf{FACT} checks whether cited evidence correctly supports the corresponding claims.
 
 \paragraph{\ding{185} Multi-Turn Refinement Extension.}
 \label{app:multiturn-data}
 To study human-in-the-loop refinement, we extend DeepResearch Bench with follow-up editing scenarios. Given an initial research query $q$ and a first-round report $y$, we introduce a subsequent user request $q'$ that asks the agent to revise the existing report rather than produce a new one from scratch. These follow-ups cover typical refinement needs, including elaborating on a specific section, adding new supporting evidence, imposing extra comparison constraints, changing the presentation style, or correcting a flawed assumption. The resulting triples $(q,y,q')$ evaluate whether an agent can satisfy the new instruction while keeping unaffected parts of the report stable. We score each revised report along five 0--20 criteria: \textbf{Ins.} measures compliance with the follow-up, \textbf{Evi.} measures support for newly added or revised claims, \textbf{Log.} measures consistency with the retained report, \textbf{Com.} measures preservation of useful original content, and \textbf{Exp.} measures clarity and rigor. The total score \textbf{Tot.} sums these five dimensions.

\paragraph{\ding{186} Interactive Decision-Making Benchmarks.}
We further include ALFWorld~\cite{ALFWorld20} and WebShop~\cite{yao2022webshop} to examine whether \mname{} can generalize from answer-oriented reasoning to interactive decision making. These benchmarks are not used as the main deep-research evaluation; instead, they serve as a complementary testbed for long-horizon state tracking, action selection, and external-environment interaction.
\begin{itemize}[leftmargin=*]
    \item \textbf{ALFWorld~\cite{ALFWorld20}.}
ALFWorld aligns TextWorld with ALFRED household tasks via shared PDDL logic, comprising over 23,000 training tasks across 6 task types and evaluation sets of 140 seen and 134 unseen tasks. Agents receive high-level instructions and textual observations, and must complete multi-step household tasks through valid actions. The benchmark tests an agent's ability to track object locations, action history, and task preconditions over extended interactions.
    \item \textbf{WebShop~\cite{yao2022webshop}.}
    WebShop is a simulated e-commerce environment built from 1.18M real-world products and 12,087 crowd-sourced shopping instructions, split into 10,587 training, 1,000 development, and 500 test instances. Given a natural-language request, an agent must issue searches, inspect product pages, compare attributes, configure options when needed, and select the item that best satisfies the user's constraints. Compared with static QA datasets, WebShop stresses grounded language understanding, query reformulation, strategic exploration, and constraint preservation during web-style interaction.
\end{itemize}

\section{Baseline Implementation Details}
\label{appendix:baselines}

% To comprehensively evaluate the effectiveness of the proposed method, 
% we compare it against a diverse set of strong baselines spanning 
% \emph{Naive}, \emph{Single-Agent}, \emph{Multi-Agent}, and 
% \emph{Agentic-RL} paradigms. 
% These baselines represent different levels of reasoning complexity, 
% coordination structure, and learning-based decision making.
% Implementation details for each category are described below.
We compare our method with baselines from four paradigms 
(\emph{Naive}, \emph{Single-Agent}, \emph{Multi-Agent}, \emph{Agentic-RL}) 
spanning different reasoning and coordination strategies. 
Implementation details for each paradigms are described below.

% \paragraph{\ding{182} Naive.}
% Naive baselines do not involve explicit agentic reasoning or coordination mechanisms. 
% They either rely solely on the LLM’s parametric knowledge or incorporate retrieval 
% in a fixed, heuristic manner.
% \begin{itemize}
%     \item \textbf{Base.}
%     A non-retrieval baseline where the LLM directly generates responses using only 
%     its internal parametric knowledge.
%     \item \textbf{FS-RAG~\cite{trivedi2023interleaving}.}
%     FS-RAG performs sentence-level retrieval, where each sentence in the input 
%     context is independently used as a query to retrieve relevant external evidence.
% \end{itemize}
\paragraph{\ding{182} Naive.} 
Naive baselines do not involve explicit agentic reasoning or coordination mechanisms. 
They either rely solely on the LLM’s parametric knowledge or incorporate retrieval 
in a fixed, heuristic manner.
\begin{itemize}
    \item \textbf{Base.} 
    A non-retrieval baseline where LLM generates answers using only its parametric knowledge.
    \item \textbf{FS-RAG~\cite{trivedi2023interleaving}.} 
    FS-RAG retrieves evidence at the sentence level, treating each input sentence independently as a query.
\end{itemize}

% \paragraph{\ding{183} Single-Agent.}
% Single-Agent baselines employ a single LLM agent that interleaves reasoning and 
% tool usage through prompting, without explicit agent-to-agent coordination.
% \begin{itemize}
%     \item \textbf{ReAct~\cite{react}.}
%     ReAct interleaves reasoning steps and action calls, enabling the model to 
%     interact with external tools such as search engines during inference.
%     \item \textbf{IRCoT~\cite{trivedi2023interleaving}.}
%     IRCoT alternates between retrieval and chain-of-thought reasoning, where 
%     intermediate reasoning steps guide retrieval and retrieved evidence informs 
%     subsequent reasoning.
% \end{itemize}

\paragraph{\ding{183} Single-Agent.} 
Single-Agent baselines use a single LLM that alternates reasoning and tool usage via prompting, without coordination between agents. 
\begin{itemize}
    \item \textbf{ReAct~\cite{react}.} 
    ReAct interleaves reasoning and action steps, allowing interaction with external tools such as search engines.
    \item \textbf{IRCoT~\cite{trivedi2023interleaving}.} 
    IRCoT alternates between retrieval and chain-of-thought reasoning, where intermediate steps guide retrieval and retrieved evidence informs subsequent reasoning.
\end{itemize}

\paragraph{\ding{184} Multi-Agent.}
Multi-Agent baselines decompose complex tasks into multiple interacting agents, 
leveraging either centralized coordination or automated agent orchestration 
strategies.
\begin{itemize}
    \item \textbf{MacNet~\cite{qianscaling}.}
    %An automated multi-agent architecture that dynamically organizes agent 
    %interactions. xxxxxx.
    % An automated multi-agent architecture that dynamically organizes agent interactions through directed acyclic graphs (DAGs), enabling scalable and efficient reasoning. It orchestrates iterative refinement among agents, enhancing collaboration robustness while mitigating context explosion and promoting emergent intelligence.
    MacNet is an automated multi-agent architecture that organizes agent interactions via directed acyclic graphs (DAGs), enabling scalable reasoning through iterative agent refinement while mitigating context explosion.
    \item \textbf{OWL~\cite{hu2025owl}.}
    % A centralized multi-agent system that employs structured agent collaboration 
    % for complex task execution. xxxxxx.
% A centralized multi-agent system that employs structured agent collaboration for complex task execution. 
% OWL decouples strategic planning from specialized execution through a modular architecture. This design enables cross-domain transferability by optimizing a domain-agnostic planner with reinforcement learning, allowing seamless adaptation to new tasks with minimal retraining.
OWL is a centralized multi-agent system that decouples high-level planning from specialized execution, using a reinforcement-learned, domain-agnostic planner to enable efficient cross-domain transfer.   
    \item \textbf{AFlow~\cite{zhangaflow}.}
    % Automated agent generation and orchestration frameworks that construct 
    % agent workflows for task completion. xxxxxx.
%      Automated agent generation and orchestration frameworks that construct 
%     agent workflows for task completion. 
% AFlow proposes an innovative automated framework that leverages Monte Carlo Tree Search (MCTS) to efficiently explore and optimize agent workflows represented as code. This method significantly reduces human effort by iteratively refining workflows through execution feedback and tree-structured experience, achieving superior performance and scalability across diverse tasks.
AFlow is an automated agent orchestration framework that employs Monte Carlo Tree Search (MCTS) to explore and optimize agent workflows represented as code through iterative execution feedback.
\end{itemize}

% \paragraph{\ding{185} Agentic-RL.}
% Agentic-RL baselines apply reinforcement learning to agentic decision making, 
% enabling models to learn when and how to invoke tools or coordinate actions 
% during multi-step reasoning.
% \begin{itemize}
%     \item \textbf{ReSearch~\cite{chen2025learning}.}
%     ReSearch applies reinforcement learning to jointly optimize reasoning and 
%     search behaviors for multi-hop question answering, without requiring 
%     supervision over intermediate reasoning steps.
%     \item \textbf{ARPO~\cite{dong2025agentic2}.}
%     ARPO introduces an entropy-aware adaptive rollout strategy that dynamically 
%     adjusts sampling at high-entropy decision points, encouraging diverse and 
%     effective tool-use behaviors.
% \end{itemize}
\paragraph{\ding{185} Agentic-RL.} 
Agentic-RL baselines use reinforcement learning to guide agentic decisions, 
learning when and how to invoke tools or coordinate actions in multi-step reasoning.
\begin{itemize}
    \item \textbf{ReSearch~\cite{chen2025learning}.} 
    ReSearch jointly optimizes reasoning and search behaviors via RL, without supervision on intermediate steps.
    \item \textbf{ARPO~\cite{dong2025agentic2}.} 
    ARPO employs an entropy-aware adaptive rollout to dynamically adjust sampling at high-uncertainty points, promoting diverse and effective tool usage.
\end{itemize}

% \paragraph{ARPO.}
% ARPO introduces an entropy-aware adaptive rollout strategy that dynamically adjusts sampling at high-entropy decision points, encouraging diverse tool-use behaviors.

% \paragraph{Mem1.}
% Mem1 is an end-to-end RL framework for long-horizon multi-turn tasks under constant memory constraints, maintaining a compact internal state for reasoning and memory consolidation.

 \paragraph{\ding{186} Deep Research Methods.}
 Deep research baselines focus on long-form report generation, where agents must iteratively search, organize evidence, and synthesize structured reports.
 \begin{itemize}
     \item \textbf{STORM~\cite{shao2024assisting}.}
     STORM builds Wikipedia-style reports by first constructing a multi-perspective outline, then collecting evidence through grounded conversations, and finally writing the report section by section.
     \item \textbf{WebWeaver~\cite{li2025webweaver}.}
     WebWeaver uses a planner-writer architecture over an evolving outline and evidence memory, enabling localized evidence retrieval during report generation.
     \item \textbf{EDR~\cite{prabhakar2025enterprise}.}
     EDR is a steerable multi-agent research system that coordinates specialized search and planning agents to produce citation-grounded reports through iterative refinement.
 \end{itemize}

\onecolumn
\section{Prompts}
\label{appendix: prompts} 

In this section, we provide a detailed introduction to the prompts used in our framework.

% ============================================================
% Center Coordinator: System Prompt
% ============================================================

\begin{tcolorbox}[
  breakable,
  enhanced,
  colback=lightgray!20,
  colframe=darkgray!80,
  title=\mname~Central Coordinator System Prompt
]
\small

\texttt{---}\\
\texttt{CURRENT\_TIME: \{\{ CURRENT\_TIME \}\}}\\
\texttt{---}

\medskip
You are an intelligent central agent responsible for managing a multi-agent system. You not only make decisions but also execute five key actions: PLAN, REFLECT, SUMMARIZE, DELEGATE, and FINISH (specific details for each action are provided below). Your role is critical for ensuring the stable operation and coordinated execution of the entire multi-agent system.

\medskip
\textbf{Current System State}
\begin{itemize}[leftmargin=*,noitemsep]
    \item \textbf{Current Node}: \texttt{\{\{current\_node\}\}}
    \item \textbf{Current Action}: \texttt{\{\{current\_action\}\}}
    \item \textbf{Memory History}:\\
    \texttt{\{\{memory\_stack\}\}}
\end{itemize}

\medskip
\texttt{\{\% if current\_action == "decision" \%\}}

\begin{itemize}[leftmargin=*,noitemsep]
    \item \textbf{Available Actions}: \texttt{\{\{available\_actions\}\}}\\
    Description:
    \begin{itemize}[leftmargin=*,noitemsep]
        \item PLAN = Reason about the current situation, analyze it, and clarify what should be done next
        \item REFLECT = Reflect on previous step and POP several no-longer-used items from the memory stack
        \item SUMMARIZE = Condense long histories
        \item DELEGATE = Assign to sub-Agent
        \item FINISH = Terminate the task only when all subtasks are completed and user requirements are fully satisfied
    \end{itemize}
    
    \item \textbf{Available Sub-Agents}: \texttt{\{\{available\_sub\_agents\}\}}\\
    (Description: \texttt{\{\{sub\_agents\_description\}\}})
\end{itemize}

\texttt{\{\% endif \%\}}

\medskip
\texttt{\{\% if current\_progress \%\}}\\
\textbf{Current Progress}: \texttt{\{\{current\_progress\}\}}\\
\texttt{\{\% endif \%\}}

\texttt{\{\% if decision\_reasoning \%\}}\\
\textbf{Decision Reasoning}: \texttt{\{\{decision\_reasoning\}\}}\\
\texttt{\{\% endif \%\}}

\texttt{\{\% if instruction \%\}}\\
\textbf{Current Instruction}: \texttt{\{\{instruction\}\}}\\
\texttt{\{\% endif \%\}}

\texttt{\{\% if summarization\_focus \%\}}\\
\textbf{Summarization Focus}: \texttt{\{\{summarization\_focus\}\}}\\
\texttt{\{\% endif \%\}}

\medskip
\texttt{\{\% if current\_action == "summarize" or current\_action == "reflect" or current\_action == "plan" \%\}}

\medskip
While the step is PLAN, SUMMARIZE, or REFLECT, provide detailed analysis in natural language format with the same language as the user query:
\begin{itemize}[leftmargin=*,noitemsep]
    \item For PLAN: Analyze the current situation comprehensively, break down complex problems, identify key factors, and develop strategic plans for next steps
    \item For REFLECT: Analyze the reflection\_target based on need\_reflect\_context, evaluate outcomes, identify issues, and suggest improvements
    \item For SUMMARIZE: Condense need\_summary\_context according to summarization\_focus, highlighting key points, patterns, and actionable insights
    \item Include specific observations, conclusions, and recommendations for next steps
    \item Maintain clarity and conciseness while preserving essential information
\end{itemize}

\texttt{\{\% endif \%\}}

\medskip
\texttt{\{\% if current\_action == "decision" \%\}}

\medskip
\textbf{Output Examples For Decision}

\medskip
If the \textbf{current action} is \textbf{Decision}, determine the next step as follows.

\medskip
\textbf{PLAN Action (Reasoning)}\\
(if the user query is en-US:)
\begin{Verbatim}[breaklines, breakanywhere]
{
  "action": "plan",
  "reasoning": "The user's query involves both technical and market analysis. Current memory stack is empty, so I need to plan the first step.",
  "params": null,
  "instruction": "Reason about the next steps based on the current state",
  "locale": "en-US"
}
\end{Verbatim}

\textbf{REFLECT Action}\\
(if the user query is en-US:)
\begin{Verbatim}[breaklines, breakanywhere]
{
  "action": "reflect",
  "reasoning": "The previous research on AI ethics trends missed recent policy updates. I should re-assign the task with refined instructions.",
  "params": null,
  "instruction": "Reflect on the previous action and its outcomes",
  "locale": "en-US"
}
\end{Verbatim}

\textbf{SUMMARIZE Action (No Parameters)}\\
(if the user query is en-US:)
\begin{Verbatim}[breaklines, breakanywhere]
{
  "action": "summarize",
  "reasoning": "The research results are extensive. Summarizing key points will help in deciding the next steps.",
  "params": null,
  "instruction": "Condense the current information into a concise summary",
  "locale": "en-US"
}
\end{Verbatim}

\textbf{DELEGATE Action (Assign Sub-Agent)}\\
(if the user query is en-US:)
\begin{Verbatim}[breaklines, breakanywhere]
{
  "action": "delegate",
  "reasoning": "I need to gather the latest market data on AI investments. The Researcher Agent is best suited for this task.",
  "params": {
    "agent_type": "researcher",
    "task_description": "Search for global AI investment trends in 2025, focusing on ethical considerations"
  },
  "instruction": "Determine which sub-Agent to assign and define the task",
  "locale": "en-US"
}
\end{Verbatim}

\begin{Verbatim}[breaklines, breakanywhere]
{
  "action": "delegate",
  "reasoning": "To further increase retrieval depth and ensure comprehensiveness and diversity, I need to use the replanner agent to formulate a specialized plan.",
  "params": {
    "agent_type": "replanner",
    "task_description": "Decompose this question into multi steps: Global AI investment trends in 2025, focusing on ethical considerations"
  }
}
\end{Verbatim}

\textbf{FINISH Action (Complete Task)}
(if the user query is en-US:)
\begin{Verbatim}[breaklines, breakanywhere]
{
  "action": "finish",
  "reasoning": "All required data has been collected, analyzed, and summarized. User's requirements have been satisfied.",
  "params": null,
  "instruction": "Task completed",
  "locale": "en-US"
}
\end{Verbatim}

\medskip

\textbf{Decision Requirements}

While the step is \textbf{decision}, you must follow these requirements and return results in JSON format with the following fields:
\begin{enumerate}[leftmargin=*]
    \item Analyze the current state and select the most appropriate action from available options.
    \item Provide a clear reasoning for the decision, justifying why the action is optimal.
    \item If choosing DELEGATE, specify the sub-Agent type and task instructions.
    \begin{itemize}[leftmargin=*,noitemsep]
        \item If choosing replanner agent: This agent can only handle \textbf{search steps planning} and is limited to decomposing retrieval tasks into actionable steps. Do not include any requirements about report writing in the task description. You MUST and ONLY use it at the beginning of the task.
    \end{itemize}
    \item Please remember to check if report is generated before you decide to FINISH the task.
    \item \textbf{You must carefully check if the current information is sufficient to support the current decision-making requirements}. Regardless of whether the information is sufficient or not, you must provide detailed reasoning. If the information is insufficient, you must take appropriate actions to supplement it (for example, by delegating to a sub-agent capable of information gathering); if the information is sufficient, you must provide detailed reasoning explaining why the current information supports the decision.
    \item \textbf{Typically, after confirming the outline, it does not mean that the current information is sufficient to cover the generation requirements}. After the outline is confirmed, you usually need to delegate a \textbf{researcher agent} to gather sufficient information to support the task fully.
    \item Return results in JSON format with the following fields:
    \begin{itemize}[leftmargin=*,noitemsep]
        \item action: Type of action (required)
        \item reasoning: Justification for the decision (required)
        \item params: Action parameters (e.g., agent\_type and task\_description for DELEGATE)
        \item instruction: Instruction corresponding to the action
        \item locale: Language of the user query (e.g., \texttt{"en-US"}, \texttt{"zh-CN"}, etc.)
    \end{itemize}
\end{enumerate}

\texttt{\{\% endif \%\}}

\medskip
\texttt{\{\% if current\_action == "plan" \%\}}

\medskip
\textbf{Output Key Points For PLAN}

\medskip
if the \textbf{current action} is \textbf{PLAN}, DO NOT give the json output, provide comprehensive reasoning and analysis in natural language format:

\medskip
\textbf{Strategic Analysis Framework}
\begin{itemize}[leftmargin=*,noitemsep]
    \item \textbf{Current Situation Assessment}: Thoroughly analyze the user query, available resources, and system state
    \item \textbf{Problem Decomposition}: Break down complex queries into manageable components and identify core objectives
    \item \textbf{Resource Evaluation}: Assess available sub-agents, tools, and information to determine optimal approach
    \item \textbf{Risk and Constraint Analysis}: Identify potential obstacles, limitations, and dependencies
    \item \textbf{Strategic Planning}: Develop a step-by-step plan with clear priorities and sequencing
\end{itemize}

\medskip
\textbf{Key Focus Areas}
\begin{itemize}[leftmargin=*,noitemsep]
    \item \textbf{Goal Clarification}: Ensure clear understanding of what needs to be accomplished
    \item \textbf{Approach Selection}: Choose the most effective methodology based on the query type and complexity
    \item \textbf{Resource Allocation}: Determine which sub-agents or tools are best suited for each task component
    \item \textbf{Timeline and Dependencies}: Consider the logical sequence of actions and any interdependencies
    \item \textbf{Success Criteria}: Define what constitutes successful completion of each planned step
\end{itemize}

\medskip
\textbf{Output Requirements}
\begin{itemize}[leftmargin=*,noitemsep]
    \item Present analysis in clear, structured format using bullet points or numbered lists
    \item Provide specific, actionable insights rather than generic observations
    \item Include concrete next steps with rationale for each recommendation
    \item Highlight critical decision points and potential alternative approaches
    \item Maintain focus on practical implementation while considering broader strategic implications
\end{itemize}

\texttt{\{\% endif \%\}}

\medskip
\texttt{\{\% if current\_action == "reflect" \%\}}

\medskip
\textbf{Output Key Points For REFLECT}

\medskip
if the \textbf{current action} is \textbf{REFLECT}, return JSON format with reflection analysis and memory cleanup decision:
\begin{Verbatim}[breaklines, breakanywhere]
{
  "analysis": "Detailed reflection analysis here",
  "pop_count": 2,
  "reasoning": "Explain why these items should be removed and what the reflection concluded"
}
\end{Verbatim}

\medskip
\textbf{Reflection Guidelines}
\begin{itemize}[leftmargin=*,noitemsep]
    \item \textbf{analysis}: Provide comprehensive reflection on the previous action
    \item \textbf{pop\_count}: Number (0 or positive integer) indicating how many recent memory stack items to remove
    \item \textbf{reasoning}: Explain the reflection conclusion and memory cleanup decision
\end{itemize}

\medskip
\textbf{Memory Stack Management Criteria}
\begin{itemize}[leftmargin=*,noitemsep]
    \item Remove duplicate or redundant information
    \item Remove outdated information that no longer applies
    \item Keep essential information supporting ongoing work
    \item Remove failed attempts or incorrect reasoning
    \item DO NOT REMOVE any history that made progress towards the final goal or decision
    \item Only remove the most recent memory stack items. Older items should not be removed unless all recent items are cleared first.
\end{itemize}

\texttt{\{\% endif \%\}}

\medskip
\texttt{\{\% if current\_action == "summarize" \%\}}

\medskip
\textbf{Output Key Points For SUMMARY}

\medskip
if the \textbf{current action} is \textbf{SUMMARIZE}, condense information based on \texttt{\{\{summarization\_focus\}\}} and \texttt{\{\{need\_summary\_context\}\}}, must meet the following requirements:
\begin{itemize}[leftmargin=*,noitemsep]
    \item \textbf{Comprehensiveness}: Ensure that all key points and critical information are included. No important content should be omitted.
    \item \textbf{Completeness}: Capture all valid inputs, core arguments, supporting data, conclusions, and recommendations from the original context.
    \item \textbf{Structured Output}: Present the summary in a clear, organized format---such as bullet points or numbered lists---to enhance readability and usability.
    \item \textbf{Information Preservation}: Even when condensing large volumes of text, prioritize distillation over omission to retain essential meaning.
    \item \textbf{Semantic Accuracy}: Maintain the original intent and meaning during summarization to avoid misinterpretation or distortion.
    \item \textbf{Highlight Key Insights}: Clearly emphasize or mark important findings, trends, and actionable recommendations (when applicable).
    \item \textbf{Contextual Relevance}: If the summary will be used in subsequent steps (e.g., decision-making or reporting), preserve logical connections to the broader context.
    \item \textbf{URL Completeness}: Ensure that ALL relevant URLs (include image URLs) are included in the summary to provide context and ensure that the summary is complete and accurate.
\end{itemize}

\texttt{\{\% endif \%\}}

\end{tcolorbox}

\medskip
% ============================================================
% Long-Term Memory Curator (Aligned with Method)
% ============================================================

\begin{tcolorbox}[
  breakable,
  enhanced,
  colback=lightgray!20,
  colframe=darkgray!80,
  title=Experience Memory Curator Prompt
]
\small

\textbf{Role}

\medskip
You are a \textbf{Experience Memory Curator}.  
Your responsibility is to maintain a structured experience memory that supports cross-task generalization by consolidating information beyond individual task executions.

\medskip
The experience memory consists of three complementary components:
\begin{itemize}[leftmargin=*,noitemsep]
  \item \textbf{User Profiles}: capture stable user attributes and preference signals.
  \item \textbf{Semantic Memory}: store factual knowledge and declarative information, particularly externally retrieved evidence.
  \item \textbf{Procedural Memory (SOPs)}: abstract key execution steps from previously completed tasks as reusable procedural patterns.
\end{itemize}

\medskip
These components are organized with a unified storage and retrieval interface.

\medskip
\textbf{Objectives}

\begin{enumerate}[leftmargin=*]
  \item Extract stable user attributes and preference signals into \texttt{user\_profiles}.
  \item Record atomic factual statements into \texttt{semantic\_memory}.
  \item Abstract reusable execution patterns into \texttt{procedural\_memory} (SOPs).
  \item Merge new information with \texttt{existing\_long\_term\_memory\_json}, preserving correctness, recency, and non-redundancy.
  \item \textbf{Return JSON only}, strictly matching the required schema.
\end{enumerate}

\medskip
\textbf{Input}

\medskip
\textbf{Task Memory:}
\begin{Verbatim}[breaklines,breakanywhere]
{{task_memory_json}}
\end{Verbatim}

\medskip
\textbf{Existing Experience Memory (can be empty):}
\begin{Verbatim}[breaklines,breakanywhere]
{{existing_long_term_memory_json}}
\end{Verbatim}

\medskip
\textbf{Current Timestamp:}
\texttt{{now\_timestamp}}

\medskip
\textbf{Output Schema (strictly required)}

\begin{Verbatim}[breaklines,breakanywhere]
{
  "user_profiles": [
    "<stable user attribute or preference signal>"
  ],
  "semantic_memory": [
    "<atomic factual statement or retrieved evidence>"
  ],
  "procedural_memory": [
    {
      "scenario": "<task context or trigger condition>",
      "procedure": "<abstracted execution steps>",
      "rationale": "<why this procedure is effective or reusable>"
    }
  ]
}
\end{Verbatim}

\medskip
\textbf{Transformation Rules}

\medskip
\textbf{User Profiles}
\begin{itemize}[leftmargin=*,noitemsep]
  \item Capture stable user attributes, preferences, and experience behavior signals.
  \item Must remain valid across tasks and sessions.
  \item Avoid task-specific, transient, or procedural details.
\end{itemize}

\medskip
\textbf{Semantic Memory}
\begin{itemize}[leftmargin=*,noitemsep]
  \item Each item is a \textbf{single factual or declarative statement}.
  \item Focus on externally retrieved or verified information when applicable.
  \item Remove duplicates or merge paraphrases.
  \item Do not include user-specific preferences or procedural knowledge.
\end{itemize}

\medskip
\textbf{Procedural Memory (SOPs)}
\begin{itemize}[leftmargin=*,noitemsep]
  \item Abstract reusable execution patterns from completed tasks.
  \item Describe \textbf{how} a task is effectively performed, not what happened in a single instance.
  \item Generalize across similar task types and contexts.
  \item Avoid time-specific or one-off execution traces.
\end{itemize}

\medskip
\textbf{Merging Behavior}
\begin{itemize}[leftmargin=*,noitemsep]
  \item Combine with \texttt{existing\_long\_term\_memory\_json}.
  \item Preserve existing entries unless they are refined or superseded by more accurate information.
  \item Append new user profile signals, semantic facts, or procedural patterns when identified.
\end{itemize}

\medskip
\textbf{Style Requirements}

\begin{itemize}[leftmargin=*,noitemsep]
  \item Write factual, neutral English.
  \item No markdown formatting, commentary, or explanations outside JSON.
  \item No internal reasoning or justification.
  \item \textbf{Output plain JSON text only.}
\end{itemize}

\end{tcolorbox}

\twocolumn
% \vspace{-1em}

\section{Case Study}

We present two representative case studies to qualitatively illustrate how the proposed framework operates under different task settings, with a particular focus on task-level memory control and cross-task experience utilization.

\paragraph{Case 1: Multi-step Medical Question Answering.}
As shown in Table~\ref{tab:case1}, the system initially issues a broad retrieval query that returns irrelevant medical content.
Instead of committing this noisy information to its internal state, the central coordinator explicitly invokes \textsc{REVISE} action and modifies the retrieval key to progressively narrow the search scope.
Through multiple iterations of retrieval, inspection, and memory revision, the system successfully identifies evidence relevant to cerebrospinal fluid pressure and arrives at the correct answer.

\begin{figure*}[t]
  \centering

  % ---------- First row ----------
  \begin{subfigure}[t]{0.5\textwidth}
    \centering
    \includegraphics[width=\linewidth]{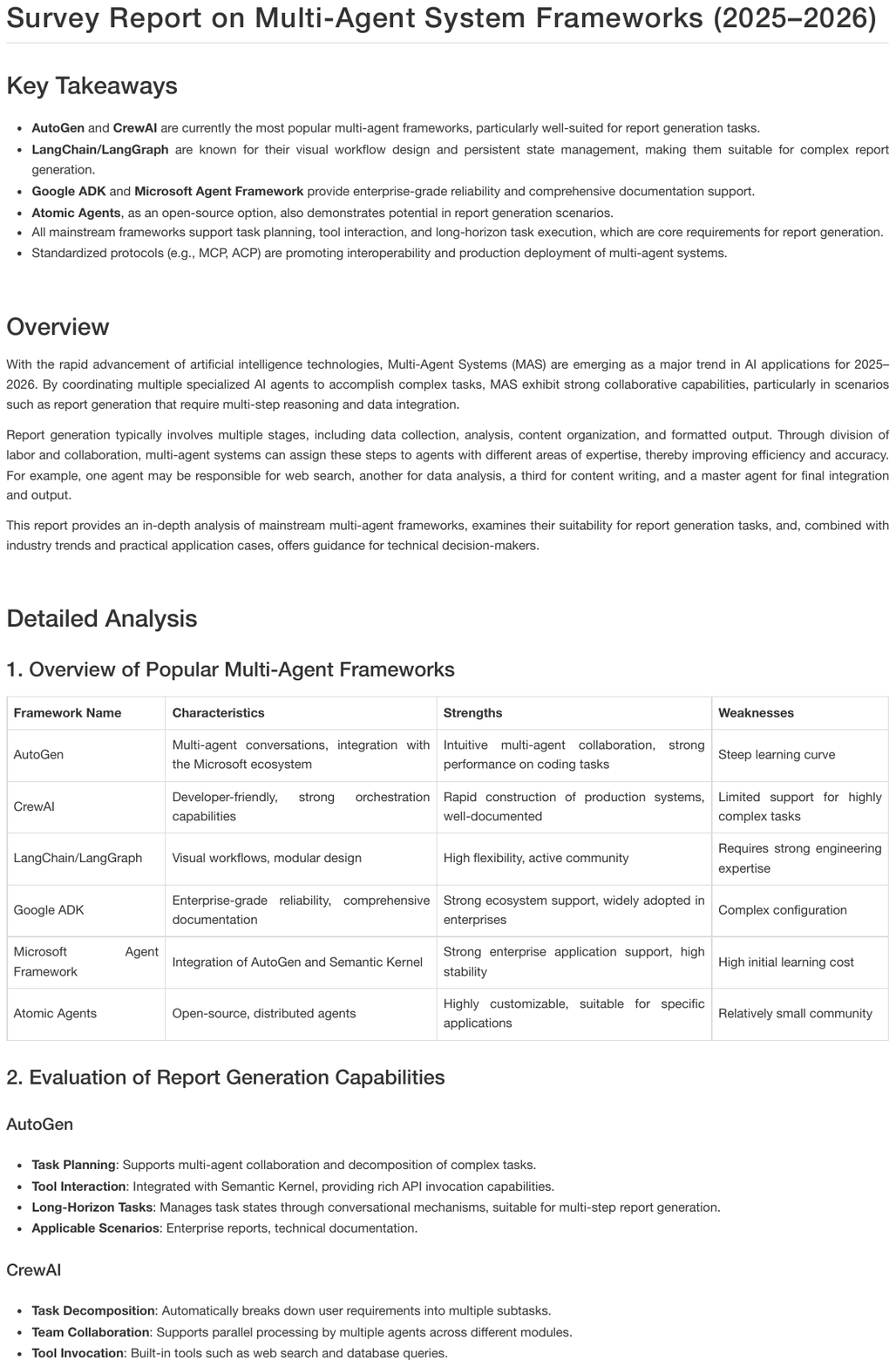}
    \caption{}
  \end{subfigure}
  \hspace{-1em}
  \begin{subfigure}[t]{0.5\textwidth}
    \centering
    \includegraphics[width=\linewidth]{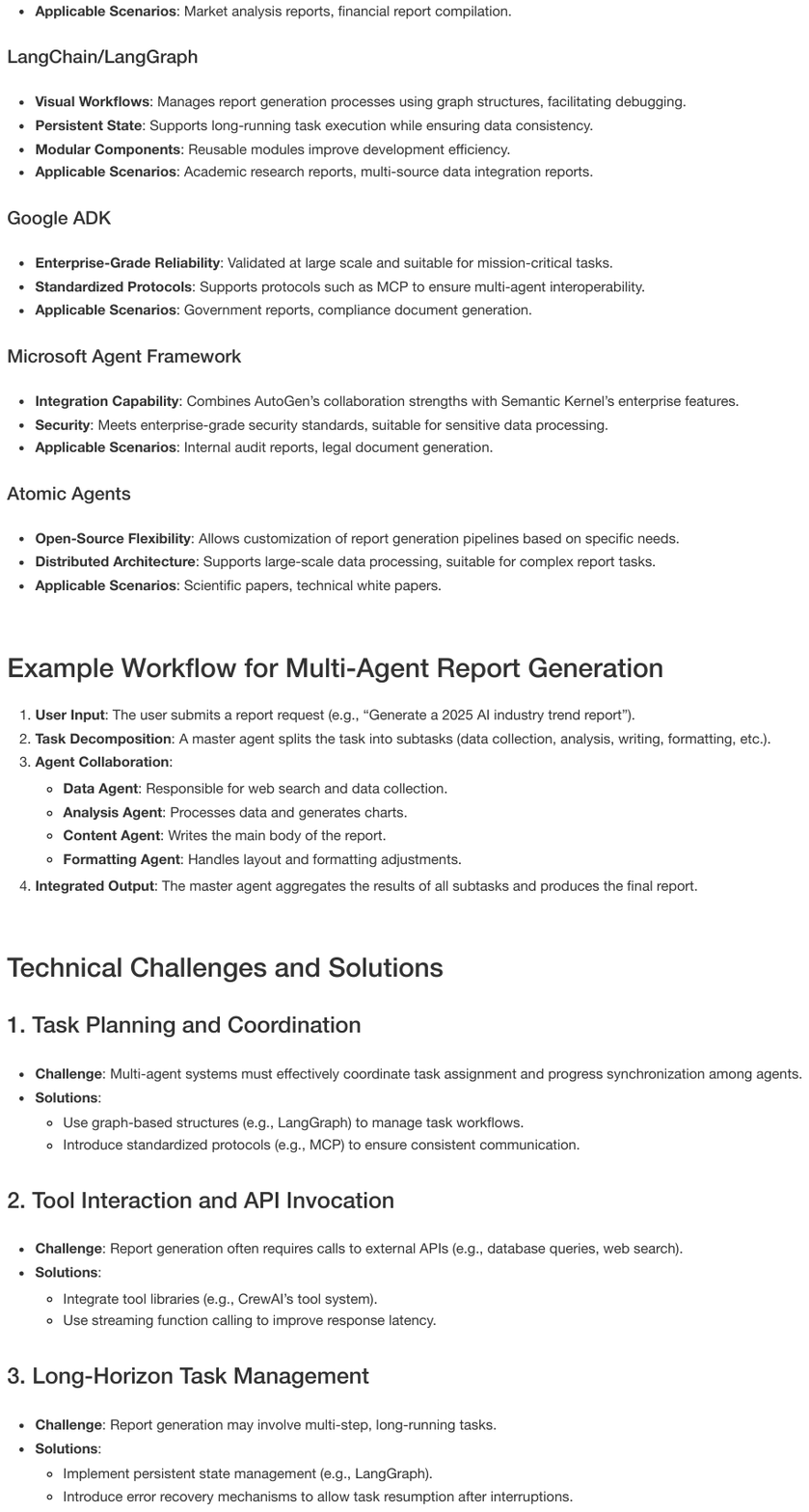}
    \caption{}
  \end{subfigure}

  \vspace{0.2em}

  % ---------- Second row ----------
  \begin{subfigure}[t]{0.5\textwidth}
    \centering
\includegraphics[
      width=\linewidth,
      trim=0 8cm 0 0,
      clip
    ]{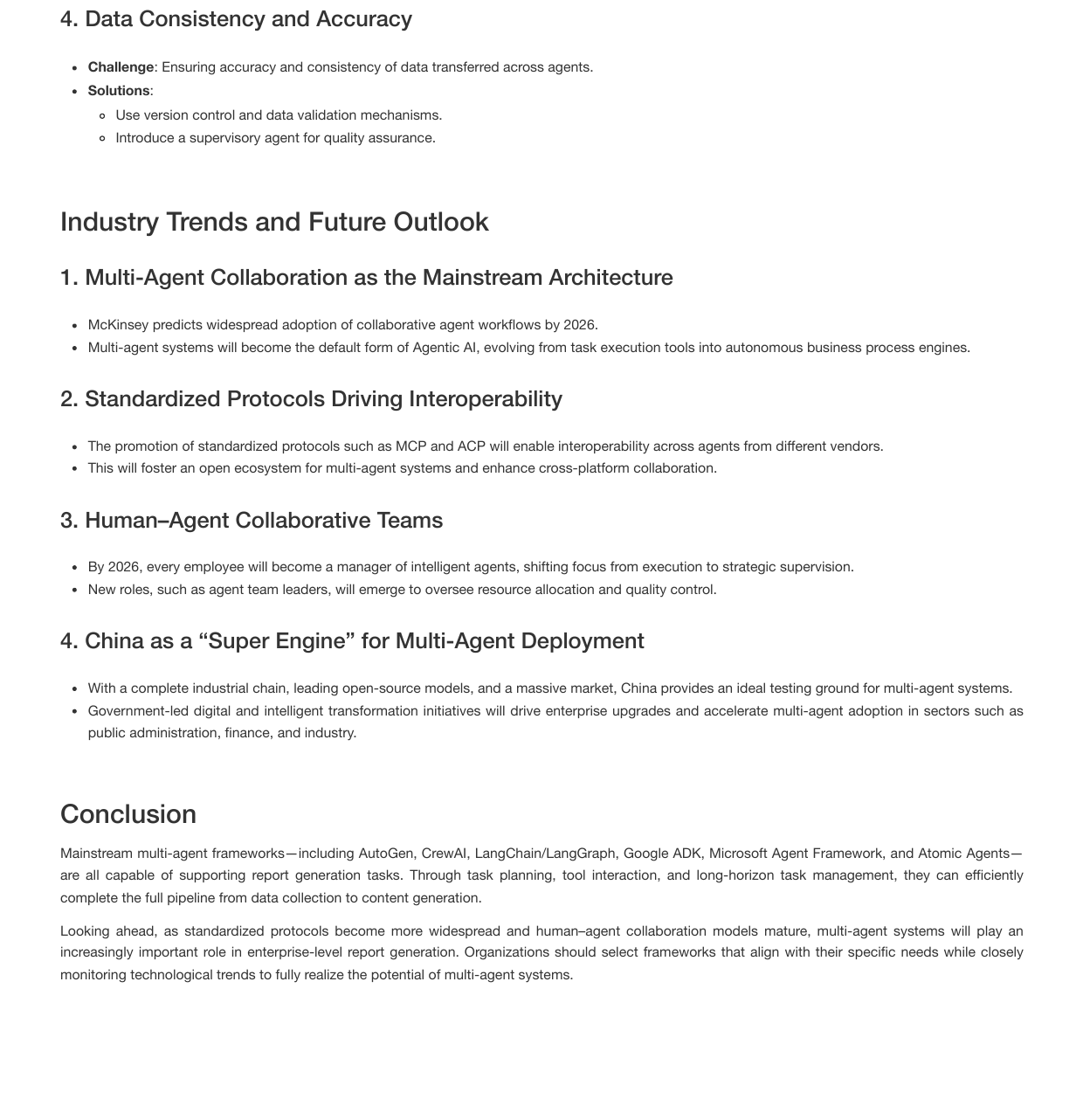}
    \caption{}
  \end{subfigure}

\caption{Case Study of \mname\ on a deepresearch task.
\textbf{Input:} ``Please summarize the recently popular multi-agent system frameworks that are capable of performing report generation tasks.''}
  \label{fig:case2}
\end{figure*}

\paragraph{Case 2: Deep Research and Report Generation.}
The second case in Figure ~\ref{fig:case2} examines a long-horizon deep research task involving open-ended information gathering and report synthesis. Through active memory management  and long-term memory storage, \mname\ retains a larger amount of high-quality, task-oriented information across extended reasoning steps, resulting in a final report that is more insightful, comprehensive, and complete.

\begin{table*}[htbp]
\centering
\begin{tabular}{
  >{\centering\arraybackslash}m{2cm}
  >{\raggedright\arraybackslash}m{11.25cm}
}
\hline
\textbf{Question} & The normal value of cerebrospinal fluid pressure for lumbar puncture in the supine position is ().\newline
A. 80-180mmH2O (0.78-1.76kPa)
B. 50-70mmH2O (0.49-0.69kPa)   \newline
C. 230-250mmH2O (2.25-2.45kPa)
D. 260-280mmH2O (2.55-2.74kPa) \\ \hline
\textbf{Ground Truth} & A. 80-180mmH2O (0.78-1.76kPa) \\ \hline
\textbf{Trajectory} & %Assistant: 
\begin{center}
\textbf{\small Central Coordinator}
\end{center}

\begin{tcolorbox}[colback=violet!12!white,colframe=white,boxsep=0pt,left=0pt,right=0pt,top=0pt,bottom=0pt,after=\vspace{-7pt} ]
% <Plan>
% The question is asking about the normal cerebrospinal fluid (CSF) pressure range during lumbar puncture in the supine position. This is a factual medical question that requires accurate clinical knowledge. Since no prior information is available in the memory stack, the next step should involve \textbf{retrieving the correct and up-to-date medical standard for CSF pressure}.
% </Plan>
<Plan>
The question is asking about the normal cerebrospinal fluid (CSF) pressure\ldots the next step should involve \textbf{retrieving the correct and up-to-date medical standard for CSF pressure}.
</Plan>
\end{tcolorbox}
\begin{tcolorbox}[colback=violet!12!white,colframe=white,boxsep=0pt,left=0pt,right=0pt,top=0pt,bottom=0pt,after=\vspace{-7pt} ]

<Delegate>
\{"handsoff":"Search Agent", "subtask":"Please find the medical standard of cerebrospinal fluid pressure"\}
</Delegate>
\end{tcolorbox}

\begin{center}
\textbf{\small Search Agent}
\end{center}

\begin{tcolorbox}[colback=green!15!white,colframe=white,boxsep=0pt,left=0pt,right=0pt,top=0pt,bottom=0pt,after=\vspace{-7pt} ]
<Search>
[Wiki\_RAG]: Cerebrospinal fluid pressure
</Search>
\end{tcolorbox}
\begin{tcolorbox}
[colback=gray!12!white,colframe=white,boxsep=0pt,left=0pt,right=0pt,top=0pt,bottom=0pt,after=\vspace{-7pt} ]
<Observation>
Observation: \{'title': 'Hydrocephalus', 'text': 'Hydrocephalus (also known as water on the brain) is a condition occurring in the brain ...'\} 
</Observation>
\end{tcolorbox}
%\\\hline \textbf{Deactivated Output (10\%)} &

\begin{center}
\textbf{\small Central Coordinator}
\end{center}

\begin{tcolorbox}[colback=violet!12!white,colframe=white,boxsep=0pt,left=0pt,right=0pt,top=0pt,bottom=0pt,after=\vspace{-7pt} ] 
<Reflect> After reviewing the information, I found that the provided content does \textbf{not directly mention the CSF pressure value} during lumbar puncture. I need to perform a new search for more \ldots%specific information. 
</Reflect>

\end{tcolorbox}

\textsc{REVISE ACTION}: Last search results have been removed

\begin{tcolorbox}[colback=violet!12!white,colframe=white,boxsep=0pt,left=0pt,right=0pt,top=0pt,bottom=0pt,after=\vspace{-7pt} ]

<Delegate>
\{"handsoff":"Search Agent", "subtask":"Search authoritative medical sources for the normal range of cerebrospinal fluid (CSF) pressure during lumbar puncture in the supine position, with a focus on standard reference values expressed in both mmH2O and kPa."\}
</Delegate>
\end{tcolorbox}

\begin{center}
\textbf{\small Search Agent}
\end{center}

\begin{tcolorbox}
[colback=green!15!white,colframe=white,boxsep=0pt,left=0pt,right=0pt,top=0pt,bottom=0pt,after=\vspace{-7pt} ]
<Search>%\newline
[Bocha\_RAG]: Lumbar puncture, cerebrospinal fluid pressure%\newline
</Search>%\newline
\end{tcolorbox}
\begin{tcolorbox}
[colback=gray!12!white,colframe=white,boxsep=0pt,left=0pt,right=0pt,top=0pt,bottom=0pt,after=\vspace{-7pt} ]
<Observation> %\newline
Observation:\{'title': 'Normal Cerebrospinal Fluid Opening Pressure', 'text': '...CSF pressure is commonly measured through a lumbar puncture...'\}
</Observation>%\newline
\end{tcolorbox}

\begin{center}
\textbf{\small Central Coordinator}
\end{center}

\begin{tcolorbox}[colback=violet!12!white,colframe=white,boxsep=0pt,left=0pt,right=0pt,top=0pt,bottom=0pt,after=\vspace{-7pt} ]
% <Summarize> % \newline
% The previous action returned authoritative medical data regarding the normal range of cerebrospinal fluid (CSF) pressure, including standard values, abnormal conditions, and underlying causes. The information from high-quality sources is summarized as follows:

% Lumbar puncture is a commonly used clinical diagnostic and therapeutic procedure \ldots \textbf{the normal CSF pressure in supine position ranges from 80 to 180 mmH2O (0.78 to 1.764 kPa)}. If the pressure exceeds 300 mmH2O (2.94 kPa), further CSF drainage should be avoided to prevent severe complications such as brain herniation \ldots
% </Summarize>
<Summarize> % \newline
The previous action returned authoritative medical data regarding the normal range of cerebrospinal fluid (CSF) pressure \ldots \textbf{the normal CSF pressure in supine position ranges from 80 to 180 mmH2O (0.78 to 1.764 kPa)}. If the pressure exceeds 300 mmH2O (2.94 kPa), further CSF drainage should be avoided to prevent severe complications such as brain herniation \ldots
</Summarize>
\end{tcolorbox}%\newline

\textsc{REVISE ACTION}: Previous search results have been summarized

\begin{tcolorbox}[colback=violet!12!white,colframe=white,boxsep=0pt,left=0pt,right=0pt,top=0pt,bottom=0pt,after=\vspace{-7pt} ]
<Delegate>
\{"handsoff":"Report Agent", "subtask":"Answer initial query based on previous observations."\}
</Delegate>
\end{tcolorbox}% \newline
\begin{center}
\textbf{\small Report Agent}
\end{center}

\begin{tcolorbox}[colback=yellow!15!white,colframe=white,boxsep=0pt,left=0pt,right=0pt,top=0pt,bottom=0pt ,after=\vspace{1pt}]%
<Conclusion> % \newline
According to authoritative medical references, the normal CSF pressure in supine position ranges from 80 to 180 mmH2O (0.78 to 1.764 kPa). Among the options, \textbf{the one closest} to this range is \textbf{option A: 80–180} mmH2O (0.78–1.76 kPa). Thus, my answer is A.  %\newline
</Conclusion>
\end{tcolorbox}
\\ \hline
\end{tabular}
\caption{ Case Study of \mname\ on a medical QA.}
\label{tab:case1}
\end{table*}

\section{Ablation Study on Memory Components}
\label{sec:ablation}

To assess the contribution of memory components, we conduct controlled ablation experiments by selectively removing task memory and experience memory while keeping all other configurations unchanged.

\begin{table*}[h]
\centering
\begin{tabular}{lcccc}
\toprule
Method & 2Wiki & MuSiQue & GAIA & FRAMES \\
\midrule
Ours & 32.92 & 16.48 & 7.71 & 16.23 \\
w/o Task Memory & 29.90 & 10.76 & 4.68 & 13.53 \\
w/o Experience Memory & 28.47 & 8.99 & 5.53 & 7.69 \\
w/o Both & 17.12 & 7.43 & 2.47 & 6.33 \\
\bottomrule
\end{tabular}
\caption{Ablation study on memory components.}
\label{ablation_memory}
\end{table*}

As shown in Table~\ref{ablation_memory}, removing task memory leads to consistent performance degradation across all benchmarks, indicating its importance for maintaining intermediate reasoning structure. Removing experience memory results in even larger drops, suggesting its critical role in cross-task generalization.

When both components are removed, performance deteriorates significantly. These results demonstrate that the observed performance gains are directly attributable to explicit memory modeling and control, rather than incidental architectural or training effects.

\section{Case Studies: Failure Modes of Experience Memory}
\label{sec:case_studies}

To better understand the behavior of StackPlanner under imperfect memory conditions, we present three representative case studies in medical consultation scenarios. Each case corresponds to a typical failure mode of experience memory: (1) noisy user-profile signals, (2) SOP overfitting, and (3) retrieval/recall errors. These cases illustrate how memory artifacts can bias coordinated decision-making, particularly under cold-start conditions.

\subsection{Case 1: Noisy User-Profile Signals}

\textbf{User Query.} 
My child is still bleeding after a tooth extraction—the gauze gets soaked quickly, and they look a bit low-energy. Should we go to the hospital now? What can I monitor at home?

\textbf{Retrieved Memory.}
(1) The user does not want to be advised to seek medical care (derived from a single emotionally charged utterance; unverified).  
(2) The user prefers concise answers (verified across multiple interactions).

\textbf{Erroneous Behavior.}
The system incorporates both profile signals into decision-making, suppressing follow-up questions and risk stratification. As a result, it produces a reassuring but incomplete response, omitting critical escalation criteria.

\textbf{Ideal Behavior.}
The system should first perform red-flag screening (e.g., bleeding severity, mental status) and provide clear escalation conditions, independent of user preference signals.

\textbf{Analysis.}
The failure arises from treating a single-source, unverified profile signal as a stable user preference. This leads to suppression of necessary medical triage steps and removal of safety-critical information.

\textbf{Recommendation.}
User profiles should primarily affect presentation-level adaptation (e.g., tone, conciseness), while high-risk decisions (e.g., triage) should enforce mandatory information collection and safety checks. Low-confidence profile entries should be down-weighted or excluded from decision-making.

\subsection{Case 2: SOP Overfitting}

\textbf{User Query.}
My child developed fever, facial swelling, and difficulty opening their mouth two days after a dental visit. Can we manage this at home?

\textbf{Retrieved Memory.}
A procedural SOP describing routine postoperative swelling management (ice + ibuprofen + observation).

\textbf{Erroneous Behavior.}
The system directly applies the SOP without checking applicability conditions, misclassifying a high-risk symptom cluster as a routine case.

\textbf{Ideal Behavior.}
The system should identify fever, progressive swelling, and trismus as red flags and recommend immediate medical evaluation.

\textbf{Analysis.}
The stored SOP lacks boundary conditions (e.g., exclusion criteria), and the system fails to perform applicability checks before execution. This results in over-generalization of procedural memory.

\textbf{Recommendation.}
SOP memory should include structured fields such as inclusion criteria, exclusion criteria, and escalation triggers. SOPs should be treated as candidate workflows rather than final decisions, with mandatory validation before execution.

\subsection{Case 3: Retrieval and Recall Errors}

\textbf{User Query.}
My child might be allergic to penicillin. The doctor prescribed amoxicillin—can they take it?

\textbf{Retrieved Memory.}
(1) General knowledge: penicillin allergy is often misreported.  
(2) Patient record: allergy ruled out (without details).  
(3) Fact: amoxicillin belongs to the penicillin class.

\textbf{Erroneous Behavior.}
The system combines incomplete patient-specific memory with general knowledge and produces a strong affirmative recommendation.

\textbf{Ideal Behavior.}
The system should verify the allergy history (reaction type, timing) and provide conservative guidance when uncertainty exists.

\textbf{Analysis.}
The critical patient-specific memory lacks essential details and provenance. The system fails to validate its completeness and over-relies on generalized knowledge.

\textbf{Recommendation.}
Critical memory entries should include provenance, timestamps, and confidence levels. When evidence is incomplete or conflicting, the system should prioritize clarification and adopt conservative decision strategies.

\paragraph{Summary.}
These cases demonstrate that memory errors can systematically bias decision-making. They highlight the necessity of explicit memory validation, structured storage, and conservative reasoning under uncertainty.

\section{System-Level Efficiency and Overhead Analysis}
\label{sec:efficiency}

To evaluate the practical efficiency of StackPlanner, we report system-level metrics including token consumption, response latency, tool usage, and interaction dynamics. Experiments are conducted on multi-hop QA and agentic benchmarks, with comparison against the ReAct baseline.

\begin{table}[h]
\centering
\begin{tabular}{lcc}
\toprule
Metric & ReAct & StackPlanner \\
\midrule
Avg. Tokens & 1597.09 & 880.89 \\
Avg. Time (s) & 30.49 & 10.15 \\
Avg. Tool Calls & 1.56 & 0.98 \\
Avg. Iterations & 14.51 & 11.38 \\
Avg. Revise Calls & 0.00 & 3.22 \\
\bottomrule
\end{tabular}
\caption{System-level efficiency comparison.}
\end{table}

StackPlanner consistently reduces token usage and response latency while requiring fewer interaction steps. Although it introduces additional REVISE operations, the overall number of iterations is lower, indicating improved decision efficiency.

This behavior aligns with training dynamics: as training progresses, the model learns to perform more effective planning and memory operations, reducing redundant reasoning and unnecessary tool usage. Overall, structured memory control enables more compact and efficient interaction trajectories.

\section{Experience Memory Capacity and Latency}
\label{sec:memory_capacity}

Different memory types are decoupled in storage and usage, allowing memory growth without affecting system execution.

Personal information memory is summarized and incorporated into the system prompt with a fixed budget (less than 512 tokens), preventing context inflation.

SOP memory and semantic memory are not directly concatenated into the prompt. Instead, they are stored in a Faiss vector database and accessed via vector retrieval. At each step, only the top-$k$ most relevant entries are retrieved and injected into the current decision context.

As a result, memory size does not directly affect prompt length. The top-$k$ mechanism ensures a bounded context size regardless of memory growth. In our experiments, the average retrieval latency is within 800 ms, which is small compared to LLM inference time and does not constitute a major system bottleneck.

\section{Cross-Domain Evaluation on Code Tasks}
\label{sec:code_eval}

To assess generalization beyond QA tasks, we conduct a preliminary evaluation on the SWE-bench-verified benchmark, which focuses on real-world code repair tasks.

As shown in Table~\ref{table:swe-bench}, despite using a smaller 7B model, StackPlanner achieves performance comparable to a significantly larger baseline. This result provides preliminary evidence that structured memory control and planning mechanisms can generalize to complex code reasoning tasks.

\section{Computational Resources and Software Environment}

% Experiments were carried out on a server equipped with two Intel Xeon Platinum 8488C processors, comprising 96 physical cores and 192 hardware threads in total. 
% The system was configured with two NVIDIA A800 GPUs, each providing 80~GB of GPU memory, and 503~GB of system RAM. 
% All experiments were conducted on Ubuntu~22.04.5~LTS.
% The software environment was based on Python~3.11.11, with package management handled through Conda (version~23.5.2). 
% Model implementation and training were performed using PyTorch~2.6.0 with CUDA support, together with HuggingFace Transformers~4.51.3 and SpaCy~3.8.4. 
% Unless otherwise specified, all models and libraries were used with their default parameter settings.
Experiments were performed on a machine running \textbf{Ubuntu 18.04.6 LTS} (\textit{bionic}), equipped with two Intel Xeon E5-2680 v4 processors providing a total of 56 logical cores (28 cores per CPU) and 377~GB of RAM. The system featured eight NVIDIA GeForce RTX 3090 GPUs, each with 24~GB of memory.
The software environment consisted of \textbf{Python 3.13.5} managed via \textbf{Conda 23.5.2}. Model implementation and training employed \textbf{PyTorch 2.9.0} with CUDA 12.2, along with HuggingFace \textbf{Transformers 4.57.1}. All software and models were used with their default configurations unless otherwise noted.
Training our method on the described hardware took roughly 45,713~seconds, while inference times varied between 40 and 300~seconds per sample depending on task complexity.

\section{The Use of Large Language Models}
% In this study, Large Language Models (LLMs) were employed as auxiliary tools to assist with language refinement and programming-related support. Their use was limited to enhancing grammatical accuracy, readability, and overall writing style, as well as offering general coding suggestions or debugging guidance. All LLM-assisted outputs were thoroughly examined and validated by the authors prior to adoption. The conception of the research, methodological design, and experiment results analyses were entirely carried out by the authors. LLMs did not contribute to the formulation of research ideas or the derivation of conclusions.
In this work, Large Language Models (LLMs) were used solely to support language polishing and programming tasks, including improving grammar, clarity, readability, and providing general coding suggestions or debugging advice. All outputs generated with LLM assistance were carefully reviewed and verified by the authors. The study's conceptualization, experimental design, and result analyses were conducted entirely by the authors, with LLMs having no role in formulating research ideas or drawing conclusions.

\end{document}